\def\isarxiv{1} 

\ifdefined\isarxiv
\documentclass[11pt]{article}
\else
\documentclass{article}

\usepackage{neurips_2023}

\fi

\usepackage{amsmath}
\usepackage{amsthm}
\usepackage{amssymb}
\usepackage{algorithm}
\usepackage{algpseudocode}
\usepackage{color}
\usepackage[english]{babel}
\usepackage{graphicx}
\usepackage{caption}
\usepackage{subcaption}
\usepackage{wrapfig,epsfig}
\usepackage{psfrag}
\usepackage{epstopdf}
\usepackage{color}
\usepackage{epstopdf}

\ifdefined\isarxiv
\usepackage[margin=1in]{geometry}
\usepackage{hyperref}
\definecolor{mydarkblue}{rgb}{0,0.08,0.45}
\hypersetup{colorlinks=true, citecolor=mydarkblue,linkcolor=mydarkblue}
\else 

\usepackage{hyperref}
\definecolor{mydarkblue}{rgb}{0,0.08,0.45}
\hypersetup{colorlinks=true, citecolor=mydarkblue,linkcolor=mydarkblue}
\fi

\graphicspath{{./figs/}}

\newtheorem{theorem}{Theorem}[section]
\newtheorem{lemma}[theorem]{Lemma}
\newtheorem{definition}[theorem]{Definition}

\newtheorem{assumption}[theorem]{Assumption}

\newtheorem{fact}[theorem]{Fact}

\newtheorem{claim}[theorem]{Claim}

\newcommand{\F}{\mathcal{F}}

\newcommand{\wh}{\widehat}
\newcommand{\wt}{\widetilde}

\newcommand{\eps}{\epsilon}

\newcommand{\abs}[1]{|#1|}

\newcommand{\supp}{\mathsf{supp}}

\newcommand{\nn}{\mathsf{nn}}
\newcommand{\ov}{\overline}

\DeclareMathOperator*{\argmin}{arg\,min}
\DeclareMathOperator*{\E}{\mathbb{E}}

\newcommand{\R}{\mathbb{R}}

\newcommand{\FF}{\mathcal{F}}

\newcommand{\midd}{\mathsf{mid}}
\renewcommand{\d}{\mathrm{d}}

\DeclareMathOperator{\tr}{tr}

\definecolor{b2}{RGB}{51,153,255}
\definecolor{mygreen}{RGB}{80,180,0}
\definecolor{yl}{RGB}{255,80,0}
\definecolor{myred}{RGB}{255, 33, 82}

\begin{document}

\ifdefined\isarxiv
\date{\empty}
    \title{Query Complexity of Active Learning for Function Family With Nearly Orthogonal Basis
    }
    \author{ 
    Xiang Chen\thanks{\texttt{xiangche@adobe.com}. Adobe Research.}
    \and 
    Zhao Song\thanks{\texttt{zsong@adobe.com}. Adobe Research.}
    \and 
    Baocheng Sun\thanks{\texttt{woafrnraetns@gmail.com}. Weizmann Institute of Science.}
    \and
    Junze Yin\thanks{\texttt{junze@bu.edu}. Boston University.}
    \and
    Danyang Zhuo\thanks{\texttt{danyang@cs.duke.edu}. Duke University.}
    }
\else

\title{Query Complexity of Active Learning for Function Family With Nearly Orthogonal Basis} 

\fi

\ifdefined\isarxiv
\begin{titlepage}
  \maketitle
  \begin{abstract}
      Many machine learning algorithms require large numbers of labeled data to deliver state-of-the-art results. In applications such as medical diagnosis and fraud detection, though there is an abundance of unlabeled data, it is costly to label the data by experts, experiments, or simulations. Active learning algorithms aim to reduce the number of required labeled data points while preserving performance. For many convex optimization problems such as linear regression and $p$-norm regression, there are theoretical bounds on the number of required labels to achieve a certain accuracy. We call this the query complexity of active learning. However, today's active learning algorithms require the underlying learned function to have an orthogonal basis. For example, when applying active learning to linear regression, the requirement is the target function is a linear composition of a set of orthogonal linear functions, and active learning can find the coefficients of these linear functions. We present a theoretical result to show that active learning does not need an orthogonal basis but rather only requires a nearly orthogonal basis. We provide the corresponding theoretical proofs for the function family of nearly orthogonal basis, and its applications associated with the algorithmically efficient active learning framework.

  \end{abstract}
  \thispagestyle{empty}
\end{titlepage}
\newpage
\else
 \maketitle
  \begin{abstract}
      Many machine learning algorithms require large numbers of labeled data to deliver state-of-the-art results. In applications such as medical diagnosis and fraud detection, though there is an abundance of unlabeled data, it is costly to label the data by experts, experiments, or simulations. Active learning algorithms aim to reduce the number of required labeled data points while preserving performance. For many convex optimization problems such as linear regression and $p$-norm regression, there are theoretical bounds on the number of required labels to achieve a certain accuracy. We call this the query complexity of active learning. However, today's active learning algorithms require the underlying learned function to have an orthogonal basis. For example, when applying active learning to linear regression, the requirement is the target function is a linear composition of a set of orthogonal linear functions, and active learning can find the coefficients of these linear functions. We present a theoretical result to show that active learning does not need an orthogonal basis but rather only requires a nearly orthogonal basis. We provide the corresponding theoretical proofs for the function family of nearly orthogonal basis, and its applications associated with the algorithmically efficient active learning framework.

  \end{abstract}
\fi

\section{Introduction}

Machine learning models can have millions to billions of parameters, and training machine learning models usually requires an abundance of labeled training data for these models to deliver state-of-art results \cite{brown2020language, devlin2018bert, radford2019language}.  
Unfortunately, in many important areas of AI, there are abundant unlabeled data but acquiring the correct label for them is costly. For example, asking expert radiologists to manually diagnose patients' medical images is more expensive than taking the medical images using increasingly cheaper imaging devices.
Another standard example is natural language processing: plenty of texts have already existed on the Internet, but labeling them requires additional human effort. 

In these domains, \textit{active learning}, where the learning algorithms can use the least number of labeled data to achieve high model accuracy, is appealing \cite{balcan2009agnostic}. In active learning, the learner has access to a set of unlabeled training data. The learner can select a subset of the unlabeled data for an \textit{oracle} to label. The choice of the subset is usually based on the characteristics of the unlabeled data, for example, uncertainty~\cite{beluch2018power,joshi2009multi,ranganathan2017deep,lewis1994sequential,seung1992query,tong2001support}, diversity~\cite{bilgic2009link,gal2017deep,guo2010active,nguyen2004active}, and expected model change~\cite{freytag2014selecting,roy2001toward,settles2007multiple}. The oracle returns the correct labels of the selected unlabeled data. After this, the learner uses the labels on the selected data to train an accurate model.

One of the most important theoretical questions in active learning is to understand how many labeled data are necessary to train a model that has high accuracy. This problem is known as the query complexity (or sample complexity) problem in active learning, and it has been well studied in the context of linear models, e.g., linear regression~\cite{cp19,BDMI13, SWZ19}. Query complexity is the key metric for the design of active learning algorithms. 

Meanwhile, there is a severe limitation of active learning: today's active learning algorithms only work when the ``learned'' function is on a strictly orthogonal basis. For example, for linear regression, the learning algorithm has to choose the coefficients for a pre-defined set of linear functions, and these functions must be orthogonal.

This raises an important \textit{theoretical} question:
\begin{center}
   {\it Can we relax this requirement of having an orthogonal basis for active learning?}
\end{center}

This question is important because this can potentially open a pathway for adopting active learning in other domains in machine learning. At the same time, this question is challenging, because breaking the assumption on the basis significantly weakens our assumptions on the basis.

In this paper, We relax the orthogonal requirement to be $\rho$-nearly orthonormal basis. We present the first active learning algorithm with a provable query complexity on a $\rho$-nearly orthonormal basis. Our algorithm is based on former research in spectral sparsification of graphs \cite{st04,ss11,BSS12,LeeSun}. 
Our main intuition is that similar to spectral sparsification for graphs, we can also view active learning as a sparsification problem: we want to get a sparse representation for a large dataset.
Our algorithm is based on the randomized BSS algorithm~\cite{LeeSun}, and we expand it to active learning on non-convex functions.

We define an iterative importance sampling procedure.  
We prove that an importance sampling procedure can sample enough labels which are sufficient for training an optimal predictor. 
So the data selected by an importance sampling procedure have the ability to recover the entire distribution. We finally find two useful iterative importance sampling procedures to complete our proof. The first one provides a bound on the size of our unlabeled dataset by i.i.d. sampling. We show that when the dataset is sufficiently large, it has the potential to approximate the original unknown distribution. Then, we introduce the second iterative importance sampling procedure which is based on randomized BSS.

This paper makes the following contributions:
\begin{itemize}
    \item We connect the spectrum sparsification theory and active learning.
    \item We proposed the first query complexity theory for a nearly orthogonal basis.
    \item We provide the first formulation of the query complexity problem for active deep learning.
    \item We introduce a new active learning algorithm for a one-hidden layer neural network. 
\end{itemize}

\paragraph{Roadmap.} 

In Section~\ref{sec:related_work}, we present the related work from three aspects: active learning, active deep learning, and theoretical active learning. In Section~\ref{sec:preli}, we explain the important notations that we use throughout this paper and introduce the fundamental definitions. In Section~\ref{sec:query_ortho}, we analyze our main result: the existence of the active learning algorithm of function family of nearly orthogonal basis. In Section~\ref{sec:query_deep}, we elucidate the other important result of this paper: the query complexity of active deep learning. In Section~\ref{sec:tech_overview}, we specifically explain the techniques we use for getting our results. In Section~\ref{sec:conclusion}, we give a conclusion for this paper.
\section{Related Work}
\label{sec:related_work}

In this section, we provide an overview of previous research related to this paper.

\paragraph{Active learning.} Active learning aims to select a few most useful data to acquire labels~\cite{balcan2009agnostic}. This allows training an accurate machine learning model and at the same time minimizes labeling costs. 
There are three types of active learning: membership query synthesis, stream-based sampling, and pool-based sampling. In membership query synthesis, the learner can query a label for any data in the input space~\cite{angluin1988queries,king2004functional}. The data does not have to be in the set of unlabeled data. Stream-based sampling means the active learning algorithm has to decide whether to query for label based on each individual data in a stream~\cite{dagan1995committee,krishnamurthy2002algorithms}. Pool-based sampling means that the active learning algorithm can access the entire set of unlabeled data and then decide which subset to query the oracle~\cite{lewis1994sequential}. 
Our algorithm is based on pool-based sampling, which is more common in practice~\cite{ren2020survey}, and our main contribution is to prove its query complexity.

\paragraph{Active deep  learning.} 
Active deep learning has recently received much attention because it is difficult to acquire labels in many domains of deep learning. One challenge of active deep learning is that deep learning inherently requires abundant labeled training data, because deep neural networks have millions to billions of parameters~\cite{brown2020language, devlin2018bert, radford2019language}. However, traditional active learning algorithms often rely on a small number of labeled data to learn. There has been a lot of effort in the research community to integrate active learning into deep learning. 
Some works focuses on improving sampling strategies~\cite{sener2017active, settles2009active, ash2019deep, gissin2019discriminative,kirsch2019batchbald,zhdanov2019diverse}, and other works target improving neural network training methods~\cite{hossain2019active,simeoni2021rethinking}. 
Our paper focuses on an important theoretical aspect of active deep learning: the query complexity, i.e., the number of labeled data required to achieve high model accuracy.

\paragraph{Theoretical active learning.}
There are many theoretical works on upper and lower bounds for learning convex functions. Perhaps the most notable of these is the sample complexity of active linear regression which has been used for sublinear algorithms for tensor decomposition \cite{diao2019optimal,fahrbach2021fast}.
A naive approach to do active linear regression is to do uniformly sampling~\cite{CDL13, HS16}. The leverage score sampling technique significantly improves the sample complexity~\cite{DMM08, MI10, Mah11, Woo14} and nearly matches the newly discovered information theoretic lower bounds \cite{cp19}. To go beyond the leverage score sampling complexity, some works~\cite{BDMI13, SWZ19} apply the deterministic linear-sample spectral sparsification method proposed by \cite{BSS12} and other works utilize a randomized version \cite{LeeSun,cp19}. Spectral sparsification techniques also can be used in attention matrix computation in large language models \cite{lsz23,dls23,dms23,gsy23_hyper,gsy23_dp}, discrepancy problems \cite{dsw22,szz22}, Fourier-signal reconstruction in compressive sensing \cite{sswz22_lattice,sswz22_quartic}.
There are also works~\cite{SM14, CKNS15} with additional assumptions or simplifications.
In addition to linear regression, the sample complexity and active learning are also considered for other convex functions~\cite{chen21d,bourgain1989approximation,musco2021active,clarkson2014sketching,ghadiri2021faster}.

\section{Preliminary}

\label{sec:preli}

\begin{algorithm}[!ht]
\caption{Our Active Learning Algorithm}\label{alg-main:ag_unknown-NN}
\begin{algorithmic}[1]
\Procedure{\textsc{ActiveLearningFramework}}{$\eps,\FF,\mathcal{X}$}
\State Let $D$ be the uniform distribution over $\mathcal{X}$
\State Generate $\{x_1,\cdots,x_k\}, \{u_1,\cdots,u_k\}\leftarrow \textsc{SelectSamples}(\FF,D,\Theta(\epsilon))$\label{step-main:good-exec}
\State Query the label $y_i$  of $x_i$ with  $ y_i \sim Y(x_i),~i\in[k]$
\State Output $\wt{f}\leftarrow\underset{f\in\FF}{\argmin}  \sum_{i=1}^k u_i\cdot |f(x_i)-y_i|^2$
\label{step-main:ERM}
\EndProcedure
\end{algorithmic}
\end{algorithm}

In this paper, we consider a regression task. We first define some notations here. In Section~\ref{sub:preli:active_learning}, we introduce the setting of active learning and explain the important steps of our algorithm. In Section~\ref{sub:preli:orthonormal}, we define the (orthonormal) basis and define the generalized version of it, namely $\rho$-nearly orthonormal basis.

\paragraph{Notations.}

For any positive integer $n$, we use $[n]$ to denote the set $\{1, 2, \cdots , n\}$. Let $\R$ represent the set containing all real numbers. 

Let $\R^{d}$ represent the set that includes all vectors with real entries, having $d$ rows and $1$ column, where $d$ is a positive integer. For all $p \in \R$, we use $\|x\|_p$ and $\|x\|_{\infty}$ to denote the $\ell_p$ and $\ell_{\infty}$ norm of the vector $x$, respectively. We use ${\cal X}$ to denote the unlabeled training set which can be viewed as a collection of vectors from $d$-dimensional space. $|{\cal X}|$ represents the size of unlabeled training dataset. We use $(D, Y)$ to denote an unknown joint distribution where the data and label are from. For any sample we draw from $(D,Y)$, we use $x_i \in \R^d$ to denote the data point and $y_i \in \R$ to denote the corresponding label. For a function $h : D \rightarrow \R$, we use $\|h(x)\|_D^2$ to denote $ \E_{x\sim D}[h^2(x)]$. For function $f$, a set of samples $S=\{x_1,\cdots, x_k\}$ and corresponding weights $w=[w_1,\cdots,w_k]$, we define $\|f(x)\|^2_{S,w}=\sum_{x_i\in S} w_i |f(x_i)|^2$. We use $O_{\epsilon}(d)$ to denote the complexity $O(d)$ when $\epsilon=\Theta(1)$.

Let $\R^{m \times n}$ represent the set that includes all matrices with real entries, having $m$ rows and $n$ columns. Here, $m$ and $n$ are positive integers.

\subsection{Active learning}
\label{sub:preli:active_learning}

In this section, we introduce the concept of active learning.

In the setting of active learning, we assume that our data $x\in\R^d$ and corresponding labels $y\in\R$ are sampled from an unknown joint distribution $(x, y)\sim (D,Y)$. More specifically, we sample data $x\sim D$ from distribution $D$ and obtain the unlabeled dataset $ \mathcal{X} \subset \R^d$.  
Since labeling is costly, we only select $k$ samples and label them according to the conditional distribution $y\sim Y(x)$.  We use our sampled data and the corresponding labels to train our neural network.

The global framework of our algorithm is shown in Algorithm \ref{alg-main:ag_unknown-NN}.
In Step \ref{step-main:good-exec}, we perform our proposed sample selecting algorithm, which selects samples $x_i,~i\in[k]$ in the dataset $\cal X$ and generate corresponding weights $u_i,~i\in[k]$. Finally, as shown in Step \ref{step-main:ERM}, we use a weighted objective to train the optimal parameter.

\subsection{\texorpdfstring{$\rho$}{}-nearly orthonormal basis}
\label{sub:preli:orthonormal}

The goal of this section is to define $\rho$-nearly orthonormal basis.

We define basis of a function family $\FF=\{ f : \R^d \rightarrow \R : f =\sum_{i=1}^d \alpha_i v_i \}$ to be a set of function $\mathcal{V}= \{v_1(x),\ldots,v_{d}(x)\}$ such that $v_i:\R^d\rightarrow\R, ~i\in[d] $ and  for any function $f\in\FF$, there exists $\alpha_1,\ldots, \alpha_{d}\in\R$ holds that $f = \alpha_1 v_1 + \ldots + \alpha_{d} v_{d}$. An orthonormal basis for distribution $D$ is a basis of function family $\FF$ with $\underset{x \sim D}{\E}[v_i(x) \cdot v_j(x)]= 1_{i=j}$. ($1_{i=j}$ denotes a binary variable that outputs $1$ if $i=j$ and outputs $0$ if $i\neq j$). However, it's hard to find an exact orthonormal basis for certain function family, such as non-linear functions. This fact motivates us to propose a generalized concept for  an orthonormal basis. Motivated by Cheap Kabatjanskii-Levenstein bound \cite{tao2019cheap} and Johnson–Lindenstrauss lemma, we define the $\rho$-nearly orthonormal basis as follows. 

\begin{definition}[$\rho$-nearly orthonormal basis]
\label{def-main:ortho-NN}
Given the distribution $D$ and desired accuracy parameter $\eps_0$, a set of  function  $\{v_1(x),\ldots,v_{{d}}(x)\}$ forms a $\rho$-nearly orthonormal basis of function family $\FF$ when the inner products taken under the distribution $D$ such that
$
    \underset{x \sim D}{\E}[v_i(x) \cdot v_j(x)]= ~1, \forall i=j \in [{d}] 
$
and
$
    | \underset{x \sim D}{\E}[v_i(x) \cdot v_j(x)] | \leq ~ \rho, \forall i \neq j \in [{d}].
$

\end{definition}

Note that this definition relies on the distribution $D$. So, the distribution should be known when we want to practically verify nearly orthonormal basis. In certain scenarios, $\rho$ can be estimated by experiments.

\section{Query Complexity of Function Family of Nearly Orthogonal Basis}
\label{sec:query_ortho}

We study the properties of the query complexity for the function family of a nearly orthogonal basis in this section.

Our main result is the following existence of an active learning algorithm of function family of nearly orthogonal basis. It states that for $\rho$-nearly orthogonal basis, we can learn an optimal predictor by taking only $O_\eps(d)$ samples to label, which is interesting because our query complexity does not depend on orthogonal parameter $\rho$, data distribution $D$, and function family $\FF$. Note that the bound on the size of the dataset is inevitable to depend on data distribution $D$, and function family $\FF$.

\begin{theorem}[Informal version of Theorem~\ref{cor:active_learning-NN}]
\label{thm:main_nearly_ortho}
If the following conditions hold
\begin{itemize}
    \item Let $\FF$ be the function family of $\rho$-nearly orthonormal basis $\{v_1(x),\ldots,v_{{d}}(x)\}$ as defined in Definition \ref{def-main:ortho-NN}.
    \item $(D, Y)$ is an unknown distribution on $(x, y)$ over $\R^d \times \R$.
    \item Let $f^*\in\FF$ minimize $\E_{(x, y)\sim (D,Y)}[\abs{f^*(x) - y}^2]$.
    \item $\rho \in (0,1/10)$ is an orthogonal parameter.
    \item $\eps \in (0,1/10)$ is an accuracy  parameter. 
    \item Let the dataset $\cal X$ contains $O((1+\rho d)(K\log (d)+ {K}/{\eps}))$ i.i.d. unlabeled samples from $D$.
    \item $K := \sup_{x \in D} \{{\sup}_{h\in\FF}\{|h(x)|^2/\|h(x)\|_D^2\}\}$ represents the condition number.
\end{itemize}
    Then, there exists a randomized algorithm that takes $O(d/{\eps})$ labels to output a $\wt{f}\in\FF$ such that, 
  \begin{align*}
     \E_{x \sim D}[|\wt{f}(x)-f^*(x)|^2] 
    \leq  O(\eps (1+\rho d^2)) \cdot \E_{(x, y)\sim (D,Y)}[|y - f^*(x)|^2].
  \end{align*}
\end{theorem}
Furthermore, when $ \rho \leq 1/d^2$, the recovery guarantee can be simplified as 
\begin{align*}
    \E_{x \sim D}[|\wt{f}(x)-f^*(x)|^2] \leq O(\eps) \cdot \E_{(x, y)\sim (D,Y)}[|y - f^*(x)|^2].    
\end{align*}

The size of unlabeled dataset depends on the distribution $D$, function family $\FF$, and desired accuracy $\eps$. Here, we use condition number $K$ to provide a bound on the size of the dataset. The dependence on $K$ is also used by \cite{cp19} in active linear regression problem. Condition number measures the concentration of the function family $\FF$ on distribution $D$. Intuitively, concentration means that there does not exist $h\in \FF$ that contains a coordinates $x\in D$ which lead to a extremely large value $h(x)$.

However, our upper bound of required labels does not depend on the data distribution $D$ or the function family $\FF$ but is determined by the dimension of the function family of nearly orthogonal basis.

To show the recover guarantee of our recovered predictor, we choose $(x, y)\sim (D,Y)$ as the test data and measure the $ \ell_2$ distance between the prediction of our predictor and the unknown optimal predictor that can see the unknown distribution $(D,Y)$. We take expectation on the error over the distribution $(D, Y)$. Moreover, $ \E_{(x, y)\sim(D,Y)}[|y - f^*(x)|^2] $ measures the expressive power of our function family of nearly orthogonal basis $\FF$.

Our result bound the perturbation between our predictor $\wt{f}$ and optimal predictor  $f^*$ within $\eps$. Another way to understand the perturbation bound is equivalently rewrite the bound  in Theorem \ref{thm-main:main} as 
\begin{align*}
 \E_{(x, y)\sim (D,Y)}[|y-\wt{f}(x)|^2] 
\leq  (1+\eps(1+\rho d^2)) \cdot \E_{(x, y)\sim (D,Y)}[|y - f^*(x)|^2]. 
\end{align*}
which shows that training with our queried labeled samples can achieve very similar prediction accuracy compared with training with unlimited labeled data.

\section{Query Complexity of Active Deep Learning}
\label{sec:query_deep}

In this paper, we also consider the query complexity of active deep learning by leveraging the active learning algorithm of function family of nearly orthogonal basis.

In Section~\ref{sub:query_deep:formulation}, we exhibit the fundamental definitions and an assumption. In Section~\ref{sub:query_deep:quadratic}, we analyze the quadratic activation and provide our result. In Section~\ref{sub:query_deep:general}, we extend the result to general activation. In Section~\ref{sub:query_deep:data}, we analyze the property of our novel data sampler.

\subsection{Problem formulation}
\label{sub:query_deep:formulation}

Our problem formulation is presented in this section.

For simplicity, we only consider two layer neural networks. Our result holds for a large class of activation including any polynomial, Sigmoid, and Swish.
First, we provide the definition of neural network. 
\begin{definition}[Two layer neural network]
\label{def-main:NN}
Let $ w_r\in\R^d$, (for each $r\in[m]$)  be the weight vector of the first layer, $ a_r\in\R$ (for each $r\in[m]$) be the output weight. We define a two layer neural network $f_\nn: \R^{d\times m} \times \R^m \times \R^d \rightarrow \R$ as the follow:
$
    f_{\nn}(W, a, x) := \frac{1}{\sqrt{m}}\sum_{r=1}^{m}a_r \phi(w_r^\top x) \in \R
$ 
where $x\in\R^{d}$ is the input, $\phi(\cdot)$ is the non-linear activation function, $d$ is the input dimension, and we use $m$ to represent the width of the neural network. Let $W:=[w_1,\cdots,w_m]\in \R^{d\times m}$  and $a := [a_1, \cdots, a_m]^\top \in \R^m$ for convenience.
\end{definition}
Similar to other works in theoretical deep learning~\cite{ll18,als19a,als19b,dllwz19, dzps19,sy19, bpsw21}, we use normalization $1/\sqrt{m}$, consider only training $W$ while fixing $a \in \{-1,+1\}^m$, and initialize each entry of $W$ to be independent sample of ${\cal N}(0,1)$. We also write $f_{\nn}(W,x) = f_{\nn}(W, a, x)$. 

Without loss of generality, we make the following assumption, which bounds the data $x\in\R^d$ in our dataset:
\begin{assumption}[Bounded samples]\label{ass_main:bounded_feature_vector}
For the distribution $D$, we have that $\max_{x\in\supp(D)} \|  x \|_2 \leq 1$. Here $\supp$ denote the support (the non-zero indices).
\end{assumption}
The assumption says that there is a bound on the maximum value of the input data. We can always assume this by rescaling the data. Inputs are bounded is a standard assumption in the optimization field~\cite{lsswy20,ll18, als19a, als19b}.

Then, we provide our formulation of the active deep  learning problem:
\begin{definition}
Given any unknown joint distribution $(D, Y)$ over $\R^d \times \R$. Let $f_\nn(W, x)$ be the neural network defined in Definition \ref{def-main:NN}. Let $W^* \in \R^{d\times m}$ minimize $\E_{(x,y)\sim (D,Y)}[\abs{f_\nn(W, x) - y}^2]$. For the dataset $\cal X$ that contains i.i.d. unlabeled samples from $D$, the goal is to design a randomized algorithm that takes $k$ labels to output $\wt{W}\in\R^{d\times  m}$ such that, 
  \begin{align*}
    \E_{x \sim D}[|f_\nn(\wt{W},x)-f_\nn(W^*,x)|^2] 
    \leq C \cdot \E_{(x, y)\sim (D,Y)}[|y - f_\nn(W^*,x)|^2],
  \end{align*}
    for some constant $C$.
\end{definition}
We give our results for two different activations, the quadratic activation and a general class of functions.

\subsection{Result for quadratic activation}
\label{sub:query_deep:quadratic}
 
In this section, we give the query complexity result for quadratic activation. Quadratic activation is widely used in theoretical studies for neural networks. The query complexity result for the quadratic function is as follows:
\begin{theorem}[ Informal version of Theorem~\ref{thm:guarantee_AL_procedure-NN}]
\label{thm-main:main_quad}
If the following conditions hold
\begin{itemize}
    \item Let $f_\nn(W, x)$ be a neural network as defined in Definition \ref{def-main:NN}.
    \item $(D, Y)$ is an unknown distribution on $(x, y)$ over $\R^d \times \R$.
    \item Let $W^* \in \R^{d\times m}$ minimizes $\E_{(x,y)\sim (D,Y)}[\abs{f_\nn(W, x) - y}^2]$.
    \item Let $0<\eps \leq O(  1/\log^3(d) )$ be an accuracy parameter.
    \item Let the dataset $\cal X$ contains $O(K\log (d )+ {K}/{\eps}) $ i.i.d. unlabeled samples from $D$.
    \item \begin{align*}
    K := \underset{x\in D}{\sup}\{ \underset{W \in \R^{d\times m}: W \neq 0}{\sup}\{ \frac{|f_\nn(W,x)|^2}{\|f_\nn(W, x)\|_D^2}\}\}.
 \end{align*}
is the condition number.
\end{itemize}

Then, there exists a randomized algorithm that takes $O({d^2}/{\eps})$ labels to output $\wt{W}\in\R^{d\times  m}$ such that
\begin{align*}
 \E_{x \sim D}[|f_\nn(\wt{W},x)-f_\nn(W^*,x)|^2] 
\leq  \eps \cdot \E_{(x, y)\sim (D,Y)}[|y - f_\nn(W^*,x)|^2].
\end{align*}

\end{theorem}
Our result for this special activation only require $O(d^2/\eps)$ labels, where $d$ is the dimension of the input data.

\subsection{Result for general activation}
\label{sub:query_deep:general}

Then we provide an upper bound on the required number of labeled samples for neural networks with general activation.
The theorem states that when the size of the dataset $|{\cal X}|$ and the number of required labels $k$ is large enough, the prediction accuracy difference is negligible between the optimal predictor $ f_\nn(W^*, x)$ trained by unlimited labeled data and our predictor $ f_\nn(\wt{W}, x)$ trained by  only $k$ labels.

\begin{theorem}[ Informal version of Theorem~\ref{thm:main}]
\label{thm-main:main}
If the following conditions hold
\begin{itemize}
    \item Let $f_\nn(W, x)$ be a neural network as defined in Definition \ref{def-main:NN}.
    \item $(D,Y)$ is an unknown distribution on $(x, y)$ over $\R^d \times \R$.
    \item Suppose that $C=(10d+\log(1/\eps_0)/\log(d))$.
    \item Suppose that the $C $-th derivative of the activation function $\phi $ of $f_\nn$ satisfied that (1) $\phi^{(C)}(x)$  exists and is continuous, and (2) $\phi^{(C)}(x)\leq 1,~x\in \R $. 
    \item Let $W^* \in \R^{d\times m}$ denote the $W$ that minimizes $\E_{(x,y)\sim (D,Y)}[\abs{f_\nn(W, x) - y}^2]$. 
    \item We define $\wh{d}:= 10d+\log(1/\eps_0)/\log(d)$.
    \item Let $\ov{d} := \binom{\wh{d} }{d}$.
    \item $\eps$ is an accuracy parameter, satisfying $0<\eps \leq O(  1/\log^3(\ov{d}) )$.
    \item $\eps_0 \in (0,1/10)$ is an additive accuracy parameter.
    \item Let the dataset $\cal X$ contains $O(K\log (\ov{d} )+ {K}/{\eps})$ i.i.d. unlabeled samples from $D$.
    \item \begin{align*}
    K := \underset{x\in D}{\sup}\{ \underset{W \in \R^{d\times m}: W \neq 0}{\sup}\{ \frac{|f_\nn(W,x)|^2}{\|f_\nn(W, x)\|_D^2}\}\}.
 \end{align*}
 is the condition number.
\end{itemize}

Then, there exists a randomized algorithm that takes $O({\ov{d}}/{\eps})$ labels to output $\wt{W}\in\R^{d\times  m}$ such that  
  \begin{align*}
     \E_{x \sim D}[|f_\nn(\wt{W},x)-f_\nn(W^*,x)|^2] 
    \leq  \eps_0 + \eps \cdot \E_{(x, y)\sim (D,Y)}[|y - f_\nn(W^*,x)|^2].
  \end{align*}
\end{theorem}

\subsection{Data sampler}
\label{sub:query_deep:data}

Treating each labeled data equally when we train a neural network would not obtain the optimal predictor because some of the data and labels are more important than others. It's natural to consider their importance in the training strategy. 

To obtain the optimal result, we combine our new query algorithm with a novel data sampler. Our theorem is as follows:
\begin{theorem}[Data Sampler]
\label{thm-main:sampler} 
If the following condition holds
\begin{itemize}
    \item Suppose our query algorithm select samples $ x_1,\cdots, x_k$ and generates weights $u_1,\cdots, u_k$.
\end{itemize}
 
 Then, the parameter $\wt{W}$ in Theorem \ref{thm-main:main} can be obtained by optimizing weighted objective as follows
\begin{align*}
    \wt{W}= {\argmin}_{W \in \R^{d\times m}} \{ \sum_{i=1}^k u_i
      \cdot |f_\nn(W, x_i)-y_i|^2
    \}.
\end{align*}
\end{theorem}
This theorem implies that we should sample our data proportional to the weight $u_i$ if we use stochastic gradient descent to train our neural network. If we sample with 
$
\Pr[x_i]=  u_i/\sum_{j=1}^{k} u_j,
$
then it is equivalent to train the weighted objective and thus, we can obtain the desired result.

\section{Overview of Techniques}
\label{sec:tech_overview}

In this section, we present the techniques that we use to support the result of this paper.

\paragraph{High-level approach}

Our goal is to find an algorithm that can recover the optimal predictor with finite data and limited labels. Thus, we have two sub-problem. We studied the upper bound of the size of the training dataset, under which the optimal predictor can be recovered. We also studied the upper bound on the number of required labeling data that can guarantee high-accuracy recovery of the ground truth.
Although those two problems seem very different, technically we handle them in the same framework. 
To see this, data in a dataset satisfy a certain distribution, thus we view the dataset $\cal X$ as a set containing all the i.i.d. samples that come from an unknown distribution $D$. If we find a general and sufficient condition that implies the high accuracy recovery guarantee, we can check whether the i.i.d sampling procedure of forming the dataset can  guarantee recovery. We can also apply the same condition to check the sample selecting procedure for the upper bound on the number of required labels.

The result says that a large explicit dataset can replace the unknown implicit distribution when we try to obtain the optimal predictor.

Now, we conclude that there are three major steps to prove our main result: we propose norm preserving condition and noise controlling condition for iterative importance sampling procedure of function family of nearly orthonormal basis. Those two conditions form a sufficient condition for a recovery guarantee. We prove that random i.i.d. sampling from an unknown distribution $D$ is an iterative importance sampling procedure that satisfied norm preserving conditions and noise controlling conditions. The result provides a bound on the size of the dataset but does not provide a bound on the number of labels. We prove that our randomized sampling procedure for the function family of $\rho$-nearly orthonormal basis is an iterative importance sampling procedure that satisfied norm preserving condition and noise controlling condition. This is our query complexity of active learning for function family of nearly orthogonal basis.

\paragraph{Iterative importance sampling procedure}
We formalize our sample selecting procedure as an iterative importance sampling procedure, which is a powerful tool to formalize a general class of sampling procedures. 

More specifically, our algorithm executes in an iterative way. In each iteration, our algorithm selects one sample from the dataset, and at the same time provides the importance weight of this sample. Our algorithm selects samples based on carefully designed distribution. The designed distribution is allowed to depend on the dataset $\cal X$ and the function family of nearly orthogonal basis $\FF$. More specifically, after each iteration, the sampling distribution changes so as to avoid the coupon-collector problem and to encourage the next sample to leverage more unknown information about the ground-truth. Interestingly, the changing of the sampling distribution does not depend on the label of any selected samples, which allows us to first decide on all the weighted labeling samples. Then, the oracle can label the selected samples in one pass.

When the iterative importance sampling procedure assigns weights to selected samples, we utilize the important sampling trick, which removes the off-match between the sampling distribution and the original data distribution of the dataset. More specifically, because each of our samples is taken from a different distribution, without reweighting, the samples can not well represent the original data distribution. So, our coefficients of sample $x$ always contain the term $\frac{D_{\cal X}(x)}{D'(x)}$, where $D_{\cal X}(x)$ is the uniform distribution over $\cal X$, $D'(x)$ is the sampling distribution $x$ selected from. Now, we can estimate the statistic measurement $h$ of an desired distribution $D_{\cal X}(x) $ by sampling in a new distribution $D'(x)$ as 
\begin{align*}
    \E_{x\sim D'} [\frac{D(x)}{D'(x)} h(x)] 
    = \int D'(x) \frac{D(x)}{D'(x)}h(x) \d x 
    =  \E_{x\sim D} [h(x)].
\end{align*}

\paragraph{Norm preserving condition}

Since we want to use a small subset of the dataset to represent the whole dataset, a critical observation is that the set of samples and their corresponding weights should preserve the norm of any function in the function family $\FF$ of $\rho$-nearly orthogonal basis, where the norm measure the average value of a function. Note that the error between our outputted function and the ground-truth is also in the function family $\FF$. If for any function in $\FF$, the function norm is preserved by samples and weights, the error over the data's unknown ground-truth distribution can be bounded by the weighted error over our samples. The latter one is explicit, thus, is much easier to bound than the error over an unknown distribution. 

Technically, let the weight 
\begin{align*}
    u_i=\beta_i \cdot D_{\cal X}(x_i) / D_i(x_i)
\end{align*}
be the corresponding weight of $x_i$, for each $i\in [k]$, where $x_i$ is sampled from $D_i$.  
We define the matrix $A$ as:
\begin{align*}
    A(i,j)=\sqrt{u_i} \cdot v_j(x_i) \in \R^{k \times {d}},
\end{align*}
where $v_1,\cdots,v_d$ is our $\rho$-nearly orthogonal basis of function family $\FF$. 
We judge the norm preserving by $\lambda(A^* A) \in [\frac{1}{2},\frac{3}{2}]$. 
To give some intuition for the eigenvalue of $A$, we claim that if $\rho$ equals to $0$, then this condition is equivalent to, for any ${h\in\FF}$, 
\begin{align*}
    {  \sum_{i=1}^k u_i \cdot |h(x_i)|^2} \in [\frac{1}{2} {\|h\|_D^2}, \frac{3}{2} {\|h\|_D^2}],
\end{align*}
which indicates that the sampling procedure preserves the mass of the signal.

\paragraph{Noise controlling condition} Norm preserving provides tools for us to upper bound the learning error between the unknown optimal predictor and our output predictor. However, norm preserving is not enough for learning the ground truth function with high accuracy.  
Although we can output function $f\in\FF$ that is optimal for optimizing $\|f(x)-y\|_{S,w}$, a good error norm in selected samples does not indicate a good error over the whole dataset. In fact, the error term contains noise and thus no longer belongs to function family $\FF$. Moreover, because we may over-fitting the label $y$ in those discrete points, the noise may be amplified outside of those selected sample points. 

This amplification of noise should be controlled carefully. This amplification of noise is determined by the nature of the function family $\FF$. Intuitively, we are hoping to use our samples to cover the information of other data. For a function $h\in\FF$ that is scaled to normalize with norm $1$, if the maximum value of $h$ does not cover by our samples, we miss it and the error becomes large. So, we should assign a larger selecting probability to potential large error coordinates. 

To mathematically define this event, we propose $\alpha$-condition number. For $\rho$-nearly orthonormal basis $\{v_1,\ldots, v_{{d}}\}$, function $h\in\FF$ and corresponding decomposition coefficient $\alpha(h)$, 
\begin{align*}
K_{\alpha, D'}=\underset{x}{\sup} \bigg\{ \frac{D(x)}{D'(x)} \cdot \underset{h \in \FF}{\sup} \big\{ \frac{|h(x)|^2}{\|\alpha(h)\|_2^2} \big\} \bigg\}. 
\end{align*}
Note that when $\rho=0$, we have $\|\alpha(h)\|_2^2=\|h\|_D^2$ and 
\begin{align*}
    \E_{x\sim D'}[\frac{D(x)}{D'(x)} \cdot   \frac{|h(x)|^2}{\|\alpha(h)\|_2^2}] 
    =  \E_{x\sim D}[  \frac{|h(x)|^2}{\|\alpha(h)\|_2^2}] 
    = \E_{x\sim D}[  \frac{|h(x)|^2}{\|h\|_D^2}] 
    =  1,
\end{align*}
which indicates that $K_{\alpha, D'}$ shows the concentration of the weighted term. For our weight 
\begin{align*}
    u_i=\beta_i \cdot D_{\cal X}(x_i) / D_i(x_i),
\end{align*}
the formal noise controlling condition is that our coefficients should be such that 
\begin{align*}
    \sum_{i=1}^k \beta_i \le \frac{3}{2}
\end{align*}
and 
\begin{align*}
    \beta_i \cdot K_{\alpha, D_i} \le \epsilon/2, ~~~~ \forall i \in [k].
\end{align*}

\paragraph{Sufficient condition for recovery guarantee}

Combine the norm preserving the condition and the noise controlling condition, we get a sufficient condition for recovering a good predictor with a good guarantee: for the optimal predictor in function family $f\in \FF$,
  \begin{align*} 
     f:=\underset{h \in \FF}{\arg\min} \underset{(x,y)\sim(D,Y)}{\E}[|y-h(x)|^2],
 \end{align*}
 the outputted $\wt{f}(x)\in\FF$, with a high  probability satisfies: 
   \begin{align*}
    \|f(x)-\wt{f}(x)\|_D^2 \le \epsilon \cdot \underset{(x,y)\sim (D,Y)}{\E}[|y-f(x)|^2].
   \end{align*}

Note that this recovery guarantee is exactly the recovery guarantee in Theorem \ref{thm:main_nearly_ortho}. So, for any sample selecting procedure, if the procedure satisfied both the norm preserving condition and the noise controlling condition, we  directly bound the number of required labeled data by the number of iterations of the sample selecting procedure. 

Moreover, by sophisticatedly applying this idea of preserving norm and controlling noise, we also get a bound on the size of the dataset $|\mathcal{X}|$.

 \paragraph{Query complexity for function
family of nearly orthogonal basis}

Here, we provide an effective procedure that can fit into the framework of norm preserving and noise controlling, and we get an upper bound of the required number of labels. This sampling procedure takes samples one by one. Each one is sampled from a distribution that depends on previously selected samples. More specifically, $D_i$ depends on $x_1,\cdots, x_{i-1}$. Our sampling procedure terminates in $O({{d}}/\eps)$ rounds, which means that the number of samples that we take is $O({{d}}/\eps)$. The key idea is that norm preserving can be viewed as a spectral sparsification problem for graphs. Our algorithm is based on the graph sparsifiers proposed in \cite{LeeSun}, which is called randomized BSS. Full details about this algorithm can be found in Appendix.

\paragraph{Bound on the size of dataset}

Here, we consider the size of the dataset. Suppose $D$ is the unknown distribution of the data, $\cal X$ consists of i.i.d random samples from $D$. For example, in the image regression task, $D$ is the unknown distribution that is determined by the world, the camera, and the preference of the photographer, while $\cal X$ is the set of images that are taken by the camera, which can be viewed as a set of i.i.d random samples from $D$. Now, we get a bound on the size of the dataset by sophisticatedly applying our iterative important sampling procedure, and our sufficient condition for recovery guarantee. To analyze the size of the dataset $|{\cal X}|$, we set our iterative important sampling procedure with $ D_i\leftarrow D$ and $ \beta_i\leftarrow 1/k$. This is exactly the process of obtaining the unlabeled dataset. 

We show that when we i.i.d. take a sufficiently large number of samples from a distribution, both the norm preserving condition and noise controlling condition naturally holds. Moreover, the number of samples required is $\Theta_{\eps} ( K_{\alpha, D'} \log ({{d}}) )$, where $\eps$ is the accuracy parameter. 
Finally, we get that, when the size of the dataset $|{\cal X}|$ is sufficiently large, the optimal predictor trained by $|{\cal X}|$  labeled samples can recover the optimal predictor trained with unlimited samples and labels. Note that this result provides the bound on the size of the dataset but does not provide the bound on the minimum number of labels.

\paragraph{Techniques specific to  active deep learning}  
To find a $\rho$-nearly orthonormal basis for neural network $f$. We do Taylor expansion on the neural network and ignore the high order terms that have order larger than $\overline{d}$. Note that, during the training process, the changes of the weight of the neural network are small. Due to the assumption bound on input data, the exponentially decreasing rate of the coefficient of Taylor expansion, and the almost invariant weight of the neural network, the cost of losing high order terms is negligible. Then, we view the first $\overline{d}$ order terms as a basis for approximating the neural network. Thus, we get that for any $W\in\R^{{d}\times m}$, there exists a function composed by our basis that can approximate $f_{\nn}(W,x)$. It remains to show that, for any function that is composed by our basis, there exists a weight $W$ for neural network $f_{\nn}(W,x)$ to approximate it, which can be proved by the expressive power of the neural network.

\section{Conclusion}
\label{sec:conclusion}

We present the first result to show that active learning does not need an orthogonal basis but rather only requires a nearly orthogonal basis. We provide a construction proof with the first active learning algorithm and corresponding theoretical proofs for the function family of nearly orthogonal basis. Moreover, we present the first active deep learning algorithm which has a provable query complexity, the minimum number of data for an oracle to label in order to maximize model accuracy. Using our algorithm, we have derived the first upper bound on the query complexity of active deep learning. Our upper bound shows that the minimum number of labeled data is determined by the smoothness of non-linear activation and the dimension of the input data. Our results have advanced the state-of-the-art query complexity analysis for convex functions to non-convex functions, such as neural networks. This work is mainly focusing on providing an upper bound. We believe showing a lower bound is also an interesting future direction. As far as we are aware, our work does not have negative societal impacts.




\newpage
\ifdefined\isarxiv

\else
 
\bibliography{ref}
\bibliographystyle{alpha}
 
\fi

\newpage


\appendix
\onecolumn
\section*{Appendix}

\paragraph{Roadmap.} 

In Section \ref{sec:app_preli}, we introduce several notations and tools that are useful in our paper. In Section \ref{sec:bound_nealy_basis}, we prove several properties of nearly orthogonal basis. In Section \ref{sec:suff_condition}, we give norm preserving condition and noise controlling condition, which suffice  high accuracy recovery. In Section \ref{sec:query_complexity}, we provide a linear sampling algorithm for a function family of nearly orthogonal basis. In Section \ref{sec:dataset}, we provide a bound for the size of the dataset for a function family of nearly orthogonal basis. In Section \ref{sec:main_nearly_app}, we provide a bound for the number of labels for the function family of nearly orthogonal basis. In Section \ref{sec:active_DL}, we provide our outcomes about active deep learning. In Section~\ref{sec:CAHouse_expriments}, details of the experiments are presented.

\section{Preliminary}\label{sec:app_preli}
In Section~\ref{sub:app_preli:notation}, we explain the meanings of the notations that we use. 
In Section~\ref{sub:app_preli:tools}, we introduce the properties and the tools we use. In Section~\ref{sub:app_preli:def}, we present some important definitions.

\subsection{Notations}\label{sub:app_preli:notation}
In this paper, we consider a regression task. We first define some notations here. 

For any positive integer $k$, we use $[k]$ to denote the set $\{1, 2, \cdots , k\}$. 

Let $\R^{d}$ represent the set that includes all vectors with real entries, having $d$ rows and $1$ column, where $d$ is a positive integer. For all $p \in \R$, we use $\|x\|_p$ and $\|x\|_{\infty}$ to denote the $\ell_p$ and $\ell_{\infty}$ norm of the vector $x$, respectively. We use $\langle x, y \rangle$ to denote the inner product of the vectors $x, y \in \R^d$, i.e., $\langle x, y \rangle = \sum_{i = 1}^d x_i y_i$. We denote unlabeled training data $x_i\in \R^d,~i\in[k_0]$, where $k_0$ denotes the number of unlabeled training data. We  also use $\mathcal{X}=(x_1,\dotsc,x_{k_0})$ to denote the unlabeled dataset. For convenience, we use $x_i\in \R^d,~i\in[k]$ to denote the labeled training data. We denote the corresponding label as $y_i\in\R, i\in[k]$. We use $(D, Y)$ to denote an unknown joint distribution where the data and label are from. For a function $h : D \rightarrow \R$, we use $\|h(x)\|_D^2$ to denote $ \E_{x\sim D}[h^2(x)]$. For function $f(x)$, a set of samples $S=\{x_1,\cdots, x_k\}$ and corresponding weights $w=[w_1,\cdots,w_k]$, we define $\|f(x)\|^2_{S,w}=\sum_{x_i\in S} w_i |f(x_i)|^2$.

Let $\R^{m \times n}$ represent the set that includes all matrices with real entries, having $m$ rows and $n$ columns. Here, $m$ and $n$ are positive integers. To denote the spectral norm of a matrix $A \in \R^{m \times n}$, we use $\|A\|$, which is defined as $\max_{\|x\|_2 = 1} \|Ax\|_2$. The transpose and the conjugate transpose of matrix $A$ are denoted as $A^\top$ and $A^*$, respectively. For all $i \in [m]$ and $j \in [n]$, we use $A(i, j)$ to denote the entry of $A$, located at the $i$-row and $j$-th column.

In the context of a matrix $A$, $\sigma_{\min}(A)$ represents the minimum singular value, and $\lambda_{\min}(A)$ represents the minimum eigenvalue. $\sigma_{\max}(A)$ represents the maximum singular value, which is equivalent to the spectral norm $|A|$. $\lambda_{\max}(A)$ denotes the maximum eigenvalue.

If we consider an arbitrary symmetric matrix $B \in \R^{n \times n}$, we say that $B$ is positive definite, namely $B \succ 0$, if, for all nonzero vectors $x \in \R^n$, $x^\top B x > 0$. For a symmetric matrix $B \in \R^{n \times n}$, if for all vectors $x \in \R^n$, $x^\top B x \geq 0$, we say that $B$ is positive semidefinite, namely $B \succeq 0$. When comparing two symmetric matrices $B$ and $C$, we say $B \succeq C$ if, for all vectors $x$, $x^\top B x$ is greater than or equal to $x^\top C x$. We use $B^{-1}$ to represent the inverse of the matrix $B$.

\subsection{Matrix Tools}
\label{sub:app_preli:tools}

We provide the matrix Chernoff bound theorem.

\begin{theorem}[Theorem 1.1 of \cite{Tropp}]\label{thm:matrix_chernoff-NN}
If the following conditions hold
\begin{itemize}
    \item $\{X_k\}$ is a finite sequence of independent, random, self-adjoint matrices of dimension ${d}$.
    \item Suppose that each random matrix satisfies
    \begin{align*}
        X_k \succeq 0 \quad \text{ and } \quad \lambda(X_k) \le R.
    \end{align*}
    \item Let $\mu_{\min} = \lambda_{\min}(\sum_k \E[X_k])$.
    \item Let $\mu_{\max}=\lambda_{\max}(\sum_k \E[X_k])$.
\end{itemize}

Then
\begin{align}
\Pr\left[ \lambda_{\min}(\sum_k X_k) \le (1-\delta)\mu_{\min}\right] &~ \le {d} \left( \frac{e^{-\delta}}{(1-\delta)^{1-\delta}}\right)^{\mu_{\min}/R} \text{ for } \delta \in [0,1], and \\
\Pr\left[\lambda_{\max}(\sum_k  X_k) \ge (1+\delta)\mu_{\max}\right] &~ \le {d} \left( \frac{e^{-\delta}}{(1+\delta)^{1+\delta}}\right)^{\mu_{\max}/R} \text{ for } \delta \ge 0 
\end{align}
\end{theorem}

The following Sherman-Morrison Formula discusses adding a rank $1$ matrix to an invertible matrix:
\begin{lemma}[Sherman-Morrison Formula]\label{lem:Sherman-Morrison}
If the following conditions hold
\begin{itemize}
    \item Let $A\in\R^{n\times n}$ be an invertible matrix.
    \item Let $u,v\in\R^n$.
    \item Suppose that $1+v^\top A^{-1}u\neq 0$.
\end{itemize}

Then it holds that
\begin{align*}
    (A+uv^\top)^{-1} = A^{-1} - \frac{A^{-1}uv^\top A^{-1}}{1+v^\top A^{-1} u}.
\end{align*}
\end{lemma}

\subsection{Definitions}
\label{sub:app_preli:def}
 
We first define the $\rho$-nearly orthonormal basis.
\begin{definition}
\label{def:ortho-NN}
We have a family of functions $\FF$ and the distribution $D$. 

We use $\{v_1,\ldots,v_{d}\}$ to represent $\FF$'s fixed $\rho$-nearly orthonormal basis. 

The inner products can be taken under the distribution $D$, namely 
\begin{align*}
    \underset{x \sim D}{\E}[v_i(x) \cdot v_j(x)]= & ~1, \forall i=j \in [d] \\ 
    \Big|\underset{x \sim D}{\E}[v_i(x) \cdot v_j(x)] \Big| \leq & ~ \rho, \forall i \neq j \in [d]
\end{align*}

Furthermore, for any function $h \in \FF$, there exists 
\begin{align*}
    \alpha(h):=(\alpha(h)_1,\dotsc,\alpha(h)_{d})
\end{align*}
under the basis $(v_1,\ldots,v_{d})$ such that 
\begin{align*}
  h(x)=\sum_{i=1}^{d} \alpha(h)_i \cdot v_i(x)
\end{align*}
\end{definition}

We define the norm of a function over a distribution.

\begin{definition}
Given distribution $D$ and $h\in\FF$, we define $\|h(x)\|_D$ as 
\begin{align*}
\|h(x)\|_D^2 := \E_{x\sim D}[ |h(x)|^2].
\end{align*}
\end{definition}

Now, we define condition number. Previous work only considers exactly orthogonal cases and we generalize it to a nearly orthogonal basis.
 
\begin{definition}
Let $D'$ be an arbitrary distribution over the domain $\R^d$.

Let $h : D \rightarrow \R$ be an arbitrary function.

Suppose 
\begin{align*}
    h^{(D')}(x)=\sqrt{\frac{D(x)}{D'(x)}} \cdot h(x)
\end{align*}
such that 
\begin{align*}
    \underset{x \sim D'}{\E} \left[ |h^{(D')}(x)|^2\right]=\underset{x \sim D'}{\E} \left[\frac{D(x)}{D'(x)} |h(x)|^2 \right]= \underset{x \sim D}{\E}\left[|h(x)|^2\right]. 
\end{align*}
When the function $ \FF$ and $D$ is clear, we use $K_{D'}$ to denote the condition number of sampling from $D'$, i.e., 
$$
K_{D'}=\underset{x}{\sup} \left\{ \underset{h \in \FF}{\sup} \left\{ \frac{|h^{(D')}(x)|^2}{\|h^{D'}\|_{D'}^2} \right\}\right\}=\underset{x}{\sup} \bigg\{ \frac{D(x)}{D'(x)} \cdot \underset{h \in \FF}{\sup} \big\{ \frac{|h(x)|^2}{\|h\|_D^2} \big\} \bigg\}.
$$
\end{definition}
We generalize the definition above to $\alpha$-condition number.

\begin{definition}
For any distribution $D'$ over the domain $\R^d$ and any function $h : D \rightarrow \R$. 
When the function $ \FF$ and $D$ is clear, we use $K_{\alpha, D'}$ to denote the $\alpha$-condition number of sampling from $D'$, i.e., 
\begin{align*}
K_{\alpha, D'}=
\underset{x}{\sup} \bigg\{ \frac{D(x)}{D'(x)} \cdot \underset{h \in \FF}{\sup} \big\{ \frac{|h(x)|^2}{\|\alpha(h)\|_2^2} \big\} \bigg\}. 
\end{align*}
\end{definition}

We define norm preserving and noise controlling conditions:

\begin{definition}
\label{def:procedure_agnostic_learning-NN}
Given a family of function $\FF$ and underlying distribution $D$, let $v_1,\ldots,v_{d}$ of $\FF$ under $D$ be defined as in Definition \ref{def:ortho-NN}. Suppose that after $k$ iterations, the random sampling procedure, $p$, will terminate and provides coefficient $\beta_i$ and a distributions $D_i$ to sample $x_i\sim D_i$ in every iteration $i \in [k]$.

We say $P$ satisfy norm preserving condition if, with probability $0.9$, the matrix $A$ with
\begin{align*}
    A(i,j)=\sqrt{u_i} \cdot v_j(x_i) \in \R^{k \times d}
\end{align*}
has 
\begin{align*}
    \lambda(A^* A) \in [\frac{1}{2},\frac{3}{2}].
\end{align*}

We say $P$ satisfies noise controlling condition if 
\begin{align*}
    \sum_{i=1}^k \beta_i \leq \frac{3}{2}
\end{align*}
and 
\begin{align*}
    \beta_i \cdot K_{\alpha, D_i} \le \epsilon/2.
\end{align*}
\end{definition}

\begin{fact}
    If the following condition holds
    \begin{itemize}
        \item For any arbitrary positive integer $d$, let $a, b \in \R^d$.
    \end{itemize}

    Then, we have
    \begin{itemize}
        \item $\langle a,b \rangle \leq \|a\|_{\infty} \cdot \|b\|_1$.
        \item $\|a\|_1 \leq \sqrt{d}\|a\|_2$.
    \end{itemize}
\end{fact}

\section{Bounds for Nearly Orthogonal Basis}\label{sec:bound_nealy_basis}
In this section, we analyze several basic properties of the function family of nearly orthogonal basis. 
In Section~\ref{sub:bound_nealy_basis:condition}, we analyze the upper and the lower bounds for the $\alpha$-condition number. In Section~\ref{sub:bound_nealy_basis:eigen}, we show that the eigenvalue of $\rho$-nearly orthonormal basis has an upper and lower bound. In Section~\ref{sub:bound_nealy_basis:norm}, we find the upper and lower bound for the function norm.

\subsection{Bound for the Condition Number}
\label{sub:bound_nealy_basis:condition}

We bound $\alpha$-condition number as follows:
 \begin{claim}
We will claim that 
\begin{align*}
 C_l K_{D'} \leq K_{\alpha, D'} \leq   C_r K_{D'}
\end{align*}
\end{claim}
\begin{proof}
Since we have that 
\begin{align*}
C_l  \frac{1}{\|h\|_D^2}
\leq  \frac{1}{\|\alpha(h)\|_2^2} 
\leq C_r \{ \frac{1}{\|h\|_D^2} \},
\end{align*}
we can claim that
\begin{align*}
C_l\cdot \underset{x}{\sup} \bigg\{ \frac{D(x)}{D'(x)} \cdot \underset{h \in \FF}{\sup} \big\{ \frac{|h(x)|^2}{\|h\|_D^2} \big\} \bigg\}\leq\underset{x}{\sup} \bigg\{ \frac{D(x)}{D'(x)} \cdot \underset{h \in \FF}{\sup} \big\{ \frac{|h(x)|^2}{\|\alpha(h)\|_2^2} \big\} \bigg\}
\end{align*}
and
\begin{align*}
\underset{x}{\sup} \bigg\{ \frac{D(x)}{D'(x)} \cdot \underset{h \in \FF}{\sup} \big\{ \frac{|h(x)|^2}{\|\alpha(h)\|_2^2} \big\} \bigg\} \leq C_r\cdot \underset{x}{\sup} \bigg\{ \frac{D(x)}{D'(x)} \cdot \underset{h \in \FF}{\sup} \big\{ \frac{|h(x)|^2}{\|h\|_D^2} \big\} \bigg\}
\end{align*}
\end{proof}
 
\subsection{Bound for the Eigenvalue}
\label{sub:bound_nealy_basis:eigen}
Here, we bound the eigenvalue of $\rho$-nearly orthonormal basis.
\begin{lemma}
\label{lem:rho-eigen}
If the following conditions hold

\begin{itemize}
    \item Let $A\in\R^{d\times d}$.
    \item Suppose $A$ satisfies
    \begin{align*}
        A(i,j) = & ~ 1,\forall i=j\in[d]\\
        |A(i,j)| \leq & ~ \rho , \forall i\neq j \in[d].
    \end{align*}
\end{itemize}

Then, we claim that 
\begin{align*}
\lambda_{\max}(A)\leq 1 +  d\rho ,\lambda_{\min}(A)\geq 1 -  d\rho.
\end{align*}
\end{lemma}
\begin{proof}

We can show that
\begin{align}\label{eq:IA_F}
\| I - A \|_F 
= & ~ ( \sum_{i=1}^d \sum_{j=1}^d (1 - A_{i,j})^2 )^{1/2} \notag\\
\leq & ~ ( d \cdot 0 + (d^2-d) \rho^2 )^{1/2} \notag\\
\leq & ~ d \rho,
\end{align}
where the first step follows from the definition of the Forbenius norm, the second step follows from our assumption from the Lemma statement, and the third step follows from simple algebra.

First, we can lower bound $\lambda_{\min}(A)$ as follows
\begin{align*}
\lambda_{\min}(A) 
\geq & ~ \lambda_{\min}(I)  - \|I-A\| \\
\geq & ~ 1 -  \|I-A\|_F \\
\geq & ~ 1 -  d\rho,
\end{align*}
where the first step follows from $\lambda_{\min}(A) \geq \lambda_{\min}(B) - \| A - B \|$, the second step follows from the fact that the eigenvalue of the identity matrix is $1$ and $\|\cdot\| \leq \|\cdot\|_F$, and the third step follows from Eq.~\eqref{eq:IA_F}.

Second, we can upper bound $\lambda_{\max}(A)$ as follows
\begin{align*}
    \lambda_{\max}(A) 
    = & ~ \|A\|\\ 
    \leq & ~ \|I\| + \|A-I\| \\
    \leq & ~ 1 +  \|A-I \|_F \\
    \leq & ~ 1 +  d\rho,
\end{align*}
where the first step follows from $\lambda_{\max}(A) = \sqrt{ \lambda_{\max}(AA^\top) } = \sigma_{\max}(A) = \|A\|$ (note that $A$ is a PSD matrix), the second step follows from the triangle inequality, the third step follows from $\|I\| = 1$ and $\|\cdot\| \leq \|\cdot\|_F$, and the last step follows from Eq.~\eqref{eq:IA_F}.

Thus, we complete the proof.
\end{proof}

\subsection{Bound for the Norm}
\label{sub:bound_nealy_basis:norm}
We bound the norm in this section.
\begin{claim}\label{cla:alpha_h}
If the following condition holds
\begin{itemize}
    \item Let $h\in\F$ be an arbitrary function.
\end{itemize}

Then, we have
\begin{align*}
    C_l\|\alpha(h)\|_2^2 
    \leq \|h\|_D^2
    \leq C_r\|\alpha(h)\|_2^2
\end{align*}
where $C_l:=1-\rho $ and $C_r:=1 +\rho(d-1)$.
\end{claim}
\begin{proof}
We can rewrite $\| h \|_{D}^2$ as follows:
\begin{align}\label{eq:rewrite_hD2}
    \E_{x \sim D} [ | \sum_{i=1}^{d} \alpha(h)_i \cdot v_i(x)|^2] 
    = & ~ \sum_{i=1}^{d} \E_{x \sim D} [   |\alpha(h)_i \cdot v_i(x)|^2] + 2\sum_{1\leq i< j\leq d} \E_{x \sim D} [   \alpha(h)_i\alpha(h)_j \cdot v_i(x)v_j(x)] \notag\\
    = & ~ \sum_{i=1}^{d} |\alpha(h)_i|^2  + 2\rho\cdot\sum_{1\leq i< j\leq d}\alpha(h)_i\alpha(h)_j  \notag\\
    = & ~ \sum_{i=1}^{d} |\alpha(h)_i|^2 - \rho \sum_{i=1}^d |\alpha(h)_i|^2 + \rho \sum_{i=1}^d |\alpha(h)_i|^2  + 2\rho\cdot\sum_{1\leq i< j\leq d}\alpha(h)_i\alpha(h)_j  \notag\\
    = & ~ (1-\rho)\|\alpha(h)\|_2^2+\rho(\sum_{i=1}^{d}\alpha(h)_i)^2,
\end{align}
where the first step follows from the property of the expectation, the second step follows from Definition~\ref{def:ortho-NN}, and the third step and last step follows from simple algebra.

We can provide an upper bound,
\begin{align*}
    \E_{x \sim D} [ | \sum_{i=1}^{d} \alpha(h)_i \cdot v_i(x)|^2]
    = & ~ (1-\rho)\|\alpha(h)\|_2^2+\rho(\sum_{i=1}^{d}\alpha(h)_i)^2\\
    \leq & ~ (1-\rho)\|\alpha(h)\|_2^2+\rho d\|\alpha(h)\|_2^2 \\
    = & ~ (1-\rho+\rho d)\|\alpha(h)\|_2^2\\
    = & ~ C_r\|\alpha(h)\|_2^2,
\end{align*}
where the first step follows from Eq.~\eqref{eq:rewrite_hD2}, the second step follows from the definition of the $\ell_2$ norm, the third step follows from simple algebra, and the last step follows from the definition of $C_r$ (see the claim statement).

We can provide a lower bound
\begin{align*}
    \E_{x \sim D} [ | \sum_{i=1}^{d} \alpha(h)_i \cdot v_i(x)|^2]
    = & ~ (1-\rho)\|\alpha(h)\|_2^2+\rho(\sum_{i=1}^{d}\alpha(h)_i)^2\\
    \geq & ~ (1-\rho)\|\alpha(h)\|_2^2\\
    = & ~ C_l\|\alpha(h)\|_2^2,
\end{align*}
where the first step follows from Eq.~\eqref{eq:rewrite_hD2}, the second step follows from $\rho \geq 0$, and the third step follows from the definition of $C_l$ (see the claim statement).

Thus, we complete the proof.
\end{proof}

\section{Sufficient Condition for Recovery Guarantee} \label{sec:suff_condition}

In this section, we prove the noise controlling condition and norm preserving condition suffice recovery guarantee. In Section~\ref{sub:suff_condition:preli}, we introduce the background to support our analysis. In Section~\ref{sub:suff_condition:rob}, we show the approximation between continuous norm and discrete norm. In Section~\ref{sub:suff_condition:main}, we provide a recovery guarantee result.

\subsection{Preliminary}
\label{sub:suff_condition:preli}

We state a tool from prior work,
\begin{lemma}[Lemma 4.3 of \cite{cp19}]
\label{lem:guarantee_dist-NN}
If the following conditions hold
\begin{itemize}
    \item Suppose that there is a random sampling procedure $P$ which will terminate after $k$ times of iterations, and the value of $k$ might not be fixed.
    \item Suppose for each iteration $i$, $P$ offers $\beta_i$, which is a coefficient, and $D_i$, which is a distribution, which sample $x_i \sim D_i$.
    \item Suppose $u_i=\beta_i \cdot \frac{D(x_i)}{D_i(x_i)}$ is the weight.
    \item There exists a matrix $A \in \R^{k \times {d}}$ such that $A(i,j)=\sqrt{u_i} \cdot v_j(x_i)$.
    \item Let $f$ be equal to 
\begin{align*}
    \underset{h \in \FF}{\arg\min} \underset{(x,y)\sim (D,Y)}{\E}[|y-h(x)|^2].
\end{align*}
\end{itemize}

Then,
\begin{align*}
\E_P\left[ \|A^*(\vec{y}_u - \vec{f}_{S,u})\|_2^2\right] \le \sup_{P} \big\{\sum_{i=1}^k \beta_i \big\} \cdot \max_j \big\{ \beta_j \cdot K_{\alpha, D_j} \big\} \underset{(x,y)\sim (D,Y)}{\E}[|y-f(x)|^2],    
\end{align*}
where $K_{\alpha, D_i}$ is the condition number for samples from $D_i$: 
\begin{align*}
    K_{\alpha, D_i}=\underset{x}{\sup} \bigg\{ \frac{D(x)}{D_i(x)} \cdot \underset{v \in \mathcal{F}}{\sup} \big\{ \frac{|v(x)|^2}{\|\alpha(v)\|_2^2} \big\} \bigg\}.
\end{align*}
\end{lemma}

\subsection{Continuous Norm vs Discrete Norm}
\label{sub:suff_condition:rob}
We prove the continuous norm can be approximated by the weighted discrete norm.
\begin{lemma}
\label{lem:operator_estimation-NN-fwd}

If the following conditions hold
\begin{itemize}
    \item Let $\eps$ be in $(0,1)$.
    \item Let $S=(x_1,\dotsc,x_k) \subset \R^{d}$, with $(u_1,\dotsc,u_k) \subset \R_{\geq 0}$ as the weights. 
    \item Suppose that $A$ is a matrix, whose dimension is $k \times d$, satisfying
    \begin{align*}
        A(i,j) := \sqrt{u_i} \cdot v_j(x_i) .
    \end{align*}
\end{itemize}

Then, 
\begin{align*}
 \lambda(A^* A) \in    [1-\eps,1+\eps].
\end{align*}
can imply that 
\begin{align*}
\|h\|^2_{S,u}:=\sum_{j=1}^k u_j \cdot |h(x_j)|^2 \in [\frac{1 -\eps}{C_r}, \frac{1+\eps}{C_l}] \cdot {\| h \|_D^2} \text{ \quad for every } h \in \FF.
\end{align*}
\end{lemma}
\begin{proof}

We have
\begin{align}\label{eq:matrix_coefficient-NN}
A \cdot \alpha(h) = \big( \sqrt{u_1} \cdot h( x_1) ,\dotsc,\sqrt{u_k} \cdot h( x_k) \big).
\end{align}
We can show
\begin{align*}
\|h\|^2_{S,u}
= & ~ \sum_{i=1}^k u_i \cdot |h( x_i)|^2 \\
= & ~ \|A \cdot \alpha(h)\|_2^2 \\
= & ~ \alpha(h)^* \cdot (A^* \cdot A) \cdot \alpha(h) \\
\in & ~ [\lambda_{\min}(A^* \cdot A),\lambda_{\max}(A^* \cdot A)] \cdot \|\alpha(h)\|_2^2 \\
\subseteq & ~ [1 -\eps, 1+\eps] \cdot \|\alpha(h)\|_2^2 \\
\subseteq & ~ [\frac{1 -\eps}{C_r}, \frac{1+\eps}{C_l}] \cdot {\| h \|_D^2}
\end{align*} 
where the first step follows from the definition of the norm, the second step follows from Eq.~\eqref{eq:matrix_coefficient-NN}, the third step follows from the definition of $\ell_2$ norm, the fourth step follows from the definition of $\lambda_{\min}$ and $\lambda_{\max}$, the fifth step follows from the Lemma statement, and the last step follows from $ C_l\|\alpha(h)\|_2^2\leq \|h\|_D^2\leq C_r \|\alpha(h)\|_2^2$ (see Claim~\ref{cla:alpha_h}). 
\end{proof}

\subsection{Recovery Guarantee}
\label{sub:suff_condition:main}

In this section, we analyze the recovery guarantee. We use the properties from Lemma~\ref{lem:guarantee_dist-NN} and Lemma~\ref{lem:operator_estimation-NN-fwd} to support our work. Our statement is as follows: 
\begin{theorem}[Formal version of Theorem~\ref{thm-main:main_quad}]
  \label{thm:guarantee_AL_procedure-NN}

  If the following conditions hold
  \begin{itemize}
      \item $\FF$ is a family of functions.
      \item Let $\eps > 0$.
      \item Let $(D,Y)$ be the joint distribution.
      \item Suppose the iterative important sampling procedure $P$ generates weights $u_i$, selects samples $x_i$, $i\in[k]$.
      \item Suppose that, for $\FF$ and $D$, sampling procedure $P$ satisfies both the norm preserving the condition and noise controlling condition in Definition~\ref{def:procedure_agnostic_learning-NN}.
      \item Let  $f=\underset{h \in \FF}{\arg\min} \underset{(x,y)\sim(D,Y)}{\E}[|y-h(x)|^2]$.
      \item Let 
   \begin{align*}
    \wt{f}=\underset{h \in \FF}{\argmin} \left\{ \sum_{i=1}^k u_i
      \cdot |h(x_i)-y_i|^2
    \right\}. 
  \end{align*}
  \end{itemize}

Then we have that with a high probability
  \begin{align*}
    \|f-\wt{f}\|_D^2 \leq \epsilon \cdot \underset{(x,y)\sim (D,Y)}{\E}[|y-f(x)|^2].
  \end{align*}
\end{theorem}
\begin{proof}

Let $S$ be 
\begin{align*}
    (x_1,\dotsc,x_k),
\end{align*}
which is a subset of $\R^{d}$, and 
\begin{align*}
    (u_1,\dotsc,u_k) \subset \R_{\geq 0}
\end{align*}
be the weights. 

$A$ is a matrix whose dimension is $k \times d$ satisfying 
\begin{align*}
  A(i,j) := \sqrt{u_i} \cdot v_j(x_i) .
\end{align*}

With an arbitrary $ h \in \FF$, let $\|h\|^2_{S,u} $ be defined as 
\begin{align*}
\|h\|^2_{S,u} := \sum_{j=1}^k u_j \cdot |h(x_j)|^2 
\end{align*}

Then, for any $ h \in \FF$
\begin{align}
A \cdot \alpha(h) = \big( \sqrt{u_1} \cdot h( x_1) ,\dotsc,\sqrt{u_k} \cdot h( x_k) \big).
\end{align}

Considering the computation of $\wt{f}$, we assume that there are weights $(u_1,\cdots,u_k)$ assigned to $(x_1,\ldots,x_k)$ and with $(y_1,\ldots,y_k)$ being the labels. 

$\vec{y}_u$ represents the vector of weighted labels 
\begin{align*}
    (\sqrt{u_1} \cdot y_1,\ldots,\sqrt{u_k} \cdot y_k).
\end{align*}
Based on Eq.~\eqref{eq:matrix_coefficient-NN}, we have that the empirical distance holds true for any $ h \in \FF$,
\begin{align*}
 \|h-y\|^2_{S,u} 
 = & ~ \sum_{i=1}^k u_i |h(x_i)-y_i|^2\\
 = & ~ \|A \cdot \alpha(h) - \vec{y}_u \|^2_{2},
\end{align*}
where the first step follows from the definition of $\|\cdot\|^2_{S,u}$ and the second step follows from Eq.~\eqref{eq:matrix_coefficient-NN} and the definition of $ \vec{y}_u$.

Let 
\begin{align*}
    \wt{f}=&~\underset{h \in \FF}{\arg\min}\big\{ \|h(x_i)-y_i\|_{S,u}\big\}\\
    =&~\underset{h \in \FF}{\arg\min}\big\{\|A \cdot \alpha(h) - \vec{y}_u \|_{2}\big\}
\end{align*}

Then, we can get
\begin{align}\label{eq:alpha_wt_f}
    \alpha(\wt{f})=(A^* \cdot A)^{-1} \cdot A^* \cdot \vec{y}_u 
\end{align}
and 
\begin{align*}
    \wt{f}=\sum_{i=1}^{d} \alpha(\wt{f})_i \cdot v_i.
\end{align*}

Let 
\begin{align*}
    f=\underset{h \in \FF}{\argmin} \big\{ \underset{(x,y) \sim (D,Y)}{\E} [|h(x)-y|^2]\big\}
\end{align*}

Consider the distance between $f$ and $\wt{f}$. For convenience, let 
\begin{align*}
    \vec{f}_u=\big(\sqrt{u_1} \cdot f(x_1),\ldots, \sqrt{u_k} \cdot f(x_k) \big).
\end{align*}
Because $f \in \FF$ and Eq.~\eqref{eq:matrix_coefficient-NN}, 
we can claim that,
\begin{align}\label{eq:alpha_f}
 \alpha(f)=(A^* \cdot A)^{-1} \cdot A^* \cdot \vec{f}_u.
\end{align}

Thus,
\begin{align*}
    \|\wt{f}-f\|_D^2 
    \in & ~ [C_l, C_r]\cdot \|\alpha(\wt{f})-\alpha(f)\|_2^2 \\
    = & ~ [C_l, C_r]\cdot\|(A^* \cdot A)^{-1} \cdot A^* \cdot (\vec{y}_u-\vec{f}_u) \|_2^2,
\end{align*}
where the first step follows from 
\begin{align*}
    \|h\|_D^2\in[C_l, C_r]\|\alpha(h)\|_2^2
\end{align*}
and
\begin{align*}
    \alpha(f-g)=\alpha(f)-\alpha(g)
\end{align*}
and the second step follows from the definition of $\alpha(\wt{f})$ and $\alpha(f)$ (see Eq~\eqref{eq:alpha_wt_f} and Eq~\eqref{eq:alpha_f}, respectively).

By Definition~\ref{def:procedure_agnostic_learning-NN}, with high constant probability, 
\begin{align*}
    \lambda(A^* \cdot A) \in [1-1/2,1+1/2].
\end{align*}

By Lemma~\ref{lem:guarantee_dist-NN}, we have that
\begin{align}\label{eq:lemma_c1}
    \E_P[\|A^* \cdot (\vec{y}_u-\vec{f}_u)\|_2^2] 
    \leq & ~ \sup_{P} \big\{\sum_{i=1}^k \beta_i \big\} \cdot \max_j \big\{ \beta_j \cdot K_{\alpha, D_j} \big\} \underset{(x,y)\sim (D,Y)}{\E}[|y-f(x)|^2] \notag \\
    \leq & ~ \frac{1}{2} \epsilon \cdot \underset{(x,y)\sim (D,Y)}{\E}[|y-f(x)|^2] 
\end{align}
where the last step follows from the choice of $\beta$.

Then, we have
\begin{align*}
  \E_{P}\big[ \|(A^* \cdot A)^{-1} \cdot A^* \cdot (\vec{y}_u-\vec{f}_u) \|_2^2 \big] 
  \leq & ~ \E_{P}\big[\lambda_{\min}(A^* \cdot A)^{-1} \cdot \| A^* \cdot (\vec{y}_u-\vec{f}_u) \|_2^2 \big] \\
  \leq & ~ 2 \cdot \E_P [ \| A^* \cdot (\vec{y}_u-\vec{f}_u) \|_2^2   ] \\
  \leq & ~ \epsilon \cdot \underset{(x,y)\sim (D,Y)}{\E}[|y-f(x)|^2],
\end{align*}
where the first step follows from definition of $\lambda_{\max}$, the second step follows from $\E_P [\lambda_{\min}(A^* \cdot A)^{-1}] \leq \E_P [\| A^* \cdot (\vec{y}_u-\vec{f}_u) \|_2^2]$, and the third step follows from Eq.~\eqref{eq:lemma_c1}.

By Markov inequality, we complete the proof. 
\end{proof}

\section{Query Complexity for Function Family of Nearly Orthogonal Basis}  
\label{sec:query_complexity}

In this section, we discuss applying the spectrum sparsification theory to an active learning algorithm. More specifically, in this section, we present a linear sampling size algorithm. In Section~\ref{sub:query_complexity:prelim}, we present the background to support our work. In Section~\ref{sub:query_complexity:analysis}, we analyze the properties of the iterative importance sampling procedure (see Algorithm~\ref{alg:BSS-NN}).

Our algorithm runs in an iterative way. In each iteration, our algorithm samples from a carefully designed distribution. In the end of each iteration, the algorithm updates its sampling distribution to improve the efficiency of sampling. 

\label{sec:app_algo_bss}
\begin{algorithm}[!ht]
\caption{Randomized Iterative Importance Sampling Procedure}\label{alg:BSS-NN} 
\begin{algorithmic}[1]
\Procedure{\textsc{SelectSamples}}{$\FF,D,\epsilon$}
\State  $\gamma\leftarrow\sqrt{\epsilon}/C_0$
\State $\midd\leftarrow ({4d/\gamma})/({1/(1-\gamma)-1/(1+\gamma)})$
\State $ B_0\leftarrow 0$
\State $l_0\leftarrow-2d/\gamma$
\State $r_0\leftarrow2d/\gamma$
\State $j \leftarrow 0$
\While {$r_{j+1}-l_{j+1}<8 {d}/\gamma$}
\State $\Phi_j \leftarrow \tr[(r_j I - B_j)^{-1}] + \tr[(B_j - l_j I)^{-1}]$ 
\State $D_j(x)\leftarrow D(x) \cdot \Big(v(x)^\top (r_j I - B_j)^{-1} v(x) + v(x)^\top (B_j - l_j I)^{-1} v(x) \Big)/\Phi_j $ 
\State Sample $x_j \sim D_j$
\State $s_j\leftarrow {\gamma}\cdot{D(x)}/  ({\Phi_j}\cdot{D_j(x)})$ 
\State $B_{j+1}\leftarrow B_j + s_j \cdot v(x_j) v(x_j)^\top$
\State $r_{j+1}\leftarrow r_j + {\gamma}/({\Phi_j (1-\gamma))}$ 
\State $l_{j+1}\leftarrow l_j + {\gamma}/({\Phi_j ( 1 +\gamma)})$
\State $j\leftarrow j+1$
\EndWhile
\State $k\leftarrow j$
\For{$j \in [k]$}
    \State $\beta_j\leftarrow {\gamma}/({\Phi_j}\cdot {\midd})$,
    \State $u_j\leftarrow  s_j/\midd$, 
\EndFor
\State \Return $x, D, u, \beta$ 
\EndProcedure
\end{algorithmic}
\end{algorithm}

\subsection{Preliminary}
\label{sub:query_complexity:prelim}

We state several tools from prior work.
\begin{lemma}[Lemma 3.3 in \cite{BSS12}]

\label{lem:eigenvalues-NN-2}
If the following condition holds
\begin{itemize}
    \item Let $j$ be an arbitrary element in $[k]$. 
\end{itemize}

Then, $\lambda(B_j)$ is an element of $(l_j,r_j)$.
\end{lemma}

\begin{lemma}[A combination of Claim 5.5 and Lemma 5.6 in \cite{cp19}]\label{clm:estimate_mid-NN}
If the following condition holds
\begin{itemize}
    \item Suppose that the while loop of the \textsc{RandomizedBSS} procedure is terminated.
\end{itemize}

Then, we get that
\begin{enumerate}
\item $r_k-l_k \le 9d/\gamma$. 
\item $(1-\frac{0.5 \gamma^2}{{d}}) \cdot \sum_{j=1}^k \frac{\gamma}{\phi_j} \le \midd \le \sum_{j=1}^k \frac{\gamma}{\phi_j}$.
\item If $\frac{r_k}{l_k} \le 1+8\gamma$, then $\lambda(A^* \cdot A) \in (1-5\gamma, 1+5 \gamma)$.
\end{enumerate}
\end{lemma}

\begin{lemma}[Lemma 3.4 in \cite{LeeSun}] 
\label{lem:bound_phi}
If the following conditions hold
\begin{itemize}
    \item Let $\eps \in (0, 1/2)$. 
    \item Suppose $w^\top (uI-A)^{-1}w\leq \eps$ 
    \item Suppose $ w^\top (A-lI)^{-1}w\leq \eps$.
\end{itemize}

Then, it holds that
\begin{align*}
    \tr[(A-lI+ww^\top)^{-1}] \leq & ~  \tr[(A-lI)^{-1}]-(1-\eps)w^\top (A-lI)^{-2}w \\
    \tr[(uI-A-ww^\top)^{-1}] \leq & ~  \tr[(uI-A)^{-1}]+(1+2\eps)w^\top (uI-A)^{-2}w.
\end{align*}
\end{lemma}

The following lemma control the potential function.
\begin{lemma}[Lemma 3.5 in \cite{LeeSun}]
\label{lem:non_increaing_potential}
It holds that 
\begin{align*}
\E_{x_j\in D_j}[\Phi_{j+1}] \leq \Phi_j.
\end{align*}
\end{lemma}

The following lemma control the randomized iterative importance sampling procedure.
\begin{lemma}[Lemma 3.7 in \cite{LeeSun}]
\label{lem:well_condition_BSS-NN-2}

If the following condition holds
\begin{itemize}
    \item Let $C$ be a constant.
\end{itemize}

Then, by greater than or equal to $0.99$ of the probability, the \textsc{RandomizedSampling} procedure can ensure 
\begin{align*}
    \frac{r_k}{l_k} \le 1+8\gamma
\end{align*}
by selecting a maximum of 
\begin{align*}
    k = C \cdot {d}/\gamma^2
\end{align*}
arbitrary points $x_1,\ldots,x_k$.

\end{lemma}

\subsection{Analysis of Iterative Importance Sampling Procedure}
\label{sub:query_complexity:analysis}

In this section, we analyze the properties of the iterative importance sampling procedure.

Our results in Lemma \ref{lem:bound_wwtop} are different than results in \cite{LeeSun}. 
First, they consider a potential function $\tr[(r_j I - B_j)^{-q}] + \tr[(B_j - l_j I)^{-q}] $ with $q\geq 10$. But, in our cases, $q=1$.  Second, their results are for the orthonormal basis but not for the $\rho$-nearly orthonormal basis.

\begin{lemma}
\label{lem:bound_wwtop}
If the following conditions hold
\begin{itemize}
    \item Let $w_i$ be defined as 
    \begin{align*}
        w_j:= \sqrt{\frac{\gamma}{v(x_j)^\top (r_j I - B_j)^{-1} v(x_j) + v(x_j)^\top (B_j - l_j I)^{-1} v(x_j)}}\cdot v(x_j).
    \end{align*}
    \item Let $\gamma \leq \min\{1/(c\log(d/\gamma'),  1/(c(1+d\rho)e^2)\}$.
\end{itemize}

Then, it holds that 
\begin{align*}
    \Pr[0\preceq w_j w_j^\top \preceq \frac{1}{c} \cdot (r_j I-B_j)]\geq&~ 1-\gamma'\\
    \Pr[0\preceq w_j w_j^\top \preceq \frac{1}{c} \cdot (B_j-l_jI)]\geq&~ 1-\gamma'
\end{align*}
\end{lemma}
\begin{proof}
Let 
\begin{align*}
    R_j = v(x_j)^\top (r_j I - B_j)^{-1} v(x_j) + v(x_j)^\top (B_j - l_j I)^{-1} v(x_j).
\end{align*}

We can claim that 
\begin{align*}
\E_{x\sim D_j}[w_j w_j^\top]
= & ~ \frac{\gamma}{\Phi_j}\E_{x\sim D}[v(x_j) v(x_j)^\top]\\
\preceq & ~ \frac{\gamma(1+d\rho)}{\Phi_j}\cdot I,
\end{align*}
where the first step follows from the definition of $w_j$ and the second step follows from simple algebra.

Let 
\begin{align}\label{eq:z_j}
    z_j=(r_j I-B_j)^{-1/2} w_j.
\end{align}

It holds that 
\begin{align*}
    \tr[z_j z_j^\top] 
    = & ~ \tr[(r_j I-B_j)^{-1/2} w_j  w_j^\top (r_j I-B_j)^{-1/2}]\\
    = & ~ \frac{\gamma}{R_j}\cdot\tr[(r_j I-B_j)^{-1/2} v_j  v_j^\top (r_j I-B_j)^{-1/2}]\\
    = & ~ \frac{\gamma}{R_j}\cdot\tr[v_j^\top (r_j I-B_j)^{-1} v_j]\\
    \leq & ~ \gamma,
\end{align*}
where the first step follows from the definition of $z_j$ (see Eq.~\eqref{eq:z_j}), the second step follows from the relationship between $w_j$ and $v_j$, the third step follows from the property of $\tr$, and the last step follows from the fact that
\begin{align*}
    \tr[v_j^\top (r_j I-B_j)^{-1} v_j] \leq R_j,
\end{align*}
and $\lambda_{\max}(z_jz_j^\top)\leq \gamma$. 

Moreover, it holds that 
\begin{align*}
    \E[z_j z_j^\top ] 
    = & ~ \frac{\gamma(1+d\rho)}{\Phi_j}\cdot (r_jI-B_j)^{-1}\\
    \preceq &~  \frac{\gamma (1+d \rho) }{\Phi_j}\cdot \lambda_{\max}(\frac{1}{r_jI-B_j}) \cdot I,
\end{align*}

This implies that
\begin{align}\label{eq:lambda_max_E}
    \lambda_{\max}(\E[z_j z_j^\top ]) \leq \frac{\gamma(1+d\rho)}{\Phi_j}\cdot \lambda_{\max}(\frac{1}{r_jI-B_j}) =: \mu
\end{align}

It holds by the Matrix Chernoff Bound (Lemma \ref{thm:matrix_chernoff-NN}) that
\begin{align*}
    \Pr[\lambda_{\max}(\E[z_j z_j^\top ]) \geq (1+\delta)\mu] \leq d \cdot (\frac{\exp(\delta)}{(1+\delta)^{(1+\delta)}})^{\mu/\gamma}.
\end{align*}

Set $1+\delta$ to be 
\begin{align}
    1+\delta
    = & ~ 1/(c\mu) \label{eq:1_delta} \\
    = & ~ \frac{\Phi_j}{c\gamma(1+d\rho)}\lambda_{\min}(r_jI-B_j) \notag\\
    \geq & ~ \frac{1}{c\gamma(1+d\rho)} \label{eq:c_gamma_1_d_rho},
\end{align}
where the second step follows from Eq.~\eqref{eq:lambda_max_E} and the third step follows from $\Phi_j\lambda_{\min}(r_jI-B_j) \geq 1$.

With probability at least
 \begin{align*}
    1-d \cdot (\frac{\exp(\delta)}{(1+\delta)^{(1+\delta)}})^{\mu/\gamma} 
    \geq & ~ 1-d \cdot (\frac{e}{1+\delta})^{\mu(1+\delta)/\gamma} \\
    = & ~ 1-d \cdot (\frac{e}{1+\delta})^{1/c\gamma} \\
    \geq & ~ 1-d \cdot (ce\gamma(1+d\rho))^{1/(c\gamma)} \\
    \geq & ~ 1-d \cdot \exp(-1/(c\gamma))\\
    \geq & ~ 1-\gamma' \\
 \end{align*}
where the first step follows from simple algebra, the second step follows from Eq.~\eqref{eq:1_delta}, the third step follows from Eq.~\eqref{eq:c_gamma_1_d_rho}, the fourth step follows from $ \gamma \leq 1/(c(1+d\rho)e^2)$, and the fifth step follows from $\gamma \leq 1/(c\log(d/\gamma'))$.

As a result, we can prove that 
\begin{align*}
     \Pr[0\preceq w_j w_j^\top \preceq \frac{1}{c} \cdot (r_j I-B_j)]\geq 1-\gamma'.
\end{align*}

Similarly, we can prove that 
\begin{align*}
    \Pr[0\preceq w_j w_j^\top \preceq \frac{1}{c} \cdot (B_j-l_jI)]\geq 1-\gamma'.
\end{align*}
\end{proof}

The proof of the following lemma is similar with Lemma 5.1 in \cite{cp19}:
\begin{lemma}

\label{lemma:BSS-NN}

    If the following conditions hold
    \begin{itemize}
        \item Given any dimension ${{d}}$ linear space $\FF$.
        \item All distribution $D$ are over the domain of $\FF$.
        \item For any $\eps>0$.
    \end{itemize}

  Then, there exists an iterative importance sampling procedure that satisfies both the norm preserving the condition and noise controlling condition in Definition \ref{def:procedure_agnostic_learning-NN}, and that terminates in $O({{d}}/\eps)$ rounds with probability $0.99$.
\end{lemma}

\section{Bound on the Size of Labelled Dataset}\label{sec:dataset}
In this section, we bound the size of the labelled dataset. To prove this result, we view the forming procedure of the dataset as an i.i.d. random sampling procedure. Then we can apply the sufficient condition of high accuracy recovery to prove the bound on the dataset.

\begin{lemma}\label{lmm:general_distribution-NN}

If the following conditions hold
\begin{itemize}
    \item Suppose that $D'$ is an arbitrary distribution over $\R^d$. 
    \item Let $C$ be a constant.
    \item Let $n$ be an arbitrary element in $\mathbb{N}^+$.
    \item Let $\epsilon$ be an arbitrary element in $(0,1)$.
    \item Let $\delta$ be an arbitrary element in $(0,1)$.
    \item $\FF$ is dimension ${d}$.
    \item Let $j$ be an arbitrary element in $[k]$.
    \item Let $k \geq \frac{C}{\eps^2} \cdot K_{D'} \log \frac{{d}}{\delta}$.
    \item Let $u_j=\frac{D(x_j)}{k \cdot D'(x_j)}$.
    \item Let $A$ be a matrix whose dimension is $k \times {d}$, where $A(i,j)=\sqrt{u_i} \cdot v_j(x_i)$.
    \item Suppose that $S=(x_1,\dotsc,x_{k})$ are independently from the distribution $D'$.
\end{itemize}

Then, $A$ satisfies
\begin{align*}
\|A^* A- I\| \le \eps,
\end{align*}
with at least $1-\delta$ of the probability.

\end{lemma}
\begin{proof}
At the same time, for any fixed $x$, 
\begin{align*}
\sum_{i \in [{d}]} |v_i^{(D')}(x)|^2  
= & ~ \underset{\alpha(h)}{\sup} \frac{|\sum_{i=1}^{d} \alpha(h)_i \cdot v^{(D')}_i(x)|^2}{\|\alpha(h)\|_2^2} \\ 
= & ~ \underset{h \in \FF}{\sup} \frac{|h^{(D')}(x)|^2}{\|\alpha(h)\|_2^2},
\end{align*}
by the tightness of the Cauchy-Schwartz inequality. 

Thus, 
\begin{align}\label{def:new_K_D}
K_{\alpha, D'} := \underset{x \in R^d}{\sup} \big\{ \underset{h \in \FF: h \neq 0}{\sup} \frac{|h^{(D')}(x)|^2}{\|\alpha(h)\|_2^2} \big\} \quad \text{ indicates } \quad \sup_{x \in R^d} \sum_{i \in [{d}]} |v_i^{(D')}(x)|^2 = K_{\alpha, D'}. 
\end{align}
For each point $x_j$ in $S$ with weight $u_j=\frac{D(x_j)}{k \cdot D'(x_j)}$, for a matrix $A$, we use $A_j$ to represent its $j$-th row.

Also, $A_j \in \R^{{d}}$ is defined by 
\begin{align}\label{eq:A(j,i)}
    A_j(i)
    = & ~ A(j,i) \notag\\
    = & ~ \sqrt{u_j} \cdot v_i(x_j) \notag\\
    = & ~ \frac{v_i^{(D')}(x_j)}{\sqrt{k}},
\end{align}
where the second step follows from the definition of $A(i, j)$ in the Lemma statement and the third step follows from the definition of $v_i^{(D')} (x_j)$.

Thus, we have
\begin{align*}
    A^* A=\sum_{j=1}^k  A_j^* \cdot A_j.
\end{align*}

For $A_j^* \cdot A_j$, it is always $\succeq 0$. 

Notice that the only non-zero eigenvalue of $A_j^* \cdot A_j$ is
\begin{align*}
\lambda(A_j^* \cdot A_j)
= & ~ A_j \cdot A_j^* \\
= & ~ \frac{1}{k} \left(\sum_{i \in [{d}]} |v_i^{(D')}(x_j)|^2 \right) \\
\leq & ~ \frac{K_{\alpha, D'}}{k},
\end{align*}
where the first step follows from simple algebra (e.g. $A_j^* A_j$ is a rank-$1$ matrix), the second step follows from Eq.~\eqref{eq:A(j,i)}, and the third step follows from Eq.~\eqref{def:new_K_D}.

At the same time, because the expectation of the entry $(i,i')$ in $A_j^* \cdot A_j$ is
\begin{align*}
\underset{x_j \sim D'}{\E}[{A(j,i)} \cdot A(j,i')]
= & ~ \underset{x_j \sim D'}{\E}[\frac{{v^{(D')}_{i}(x_j)} \cdot v^{(D')}_{i'}(x_j)}{k}]\\
= & ~ \underset{x_j \sim D'}{\E}[\frac{D(x) \cdot {v_{i}(x_j)} \cdot v_{i'}(x_j)}{k \cdot D'(x_j) }]\\
= & ~ \underset{x_j \sim D}{\E}[\frac{{v_{i}(x_j)} \cdot v_{i'}(x_j)}{k}],
\end{align*}
where the first step follows from Eq.~\eqref{eq:A(j,i)}, and the second step follows from definition of $v_i^{(D')} (x_j)$, and the third step follows from definition of expectation.

We can claim that 
\begin{align*}
\underset{x_j \sim D'}{\E}[{A(j,i)} \cdot A(j,i')] = & ~1/k, \forall i=i' ,\\ 
\underset{x_j \sim D'}{\E}[{A(j,i)} \cdot A(j,i')]  \leq & ~ \rho/k, \forall i \neq i'.
\end{align*}
As a result, we have that 
\begin{align*}
\lambda_{\min}(\sum_{j=1}^k \E[A_j^* \cdot A_j]) \geq & ~ 1-\rho d,\\
\lambda_{\max}(\sum_{j=1}^k \E[A_j^* \cdot A_j]) \leq & ~ 1+\rho d.
\end{align*}

Now we apply Theorem~\ref{thm:matrix_chernoff-NN} for 
\begin{align*}
    A^* A=\sum_{j=1}^k (A_j^* \cdot A_j),
\end{align*}
so we have
\begin{align*}
& ~ \Pr\left[\lambda(A^* A) \notin [(1-\eps)(1-\rho  d),(1+\eps)(1+\rho d)]\right] \\
\leq & ~ {d} \left( \frac{e^{-\eps}}{(1-\eps)^{1-\eps}}\right)^{(1-\rho d)/\frac{K_{\alpha, D'}}{k}} + {d} \left( \frac{e^{-\eps}}{(1+\eps)^{1+\eps}}\right)^{(1-\rho  d)/\frac{K_{\alpha, D'}}{k}}\\
\leq & ~ 2d \cdot \exp({-\frac{\eps^2 \cdot \frac{k(1-\rho d)}{K_{\alpha, D'}}}{3}} )\\
\leq & ~ \delta,
\end{align*}
where the first step follows from concentration inequality, the second step follows from simple algebra, and the last step follows from 
\begin{align*}
k \geq \frac{6 K_{\alpha, D'} \log ( {{d}}/{\delta} ) }{\eps^2(1-\rho d)}.
\end{align*}

Thus, we complete the proof.
\end{proof}

The following lemma bound the size of the labeled dataset.
\begin{lemma}\label{lem:agnostic_learning_single_distribution-NN}

If the following conditions hold
\begin{itemize}
    \item Given any distribution $D'$ with the same support of $D$
    \item Let $\epsilon > 0$.
\end{itemize}

 Then, the random sampling procedure with 
 \begin{align*}
     k=\Theta(K_{\alpha, D'} \log {d} + \frac{K_{D'}}{\eps})
 \end{align*}
 i.i.d.~random samples from $D'$ and coefficients 
 \begin{align*}
     \beta_i=1/k ,\forall i \in [k]
 \end{align*}
 is an iterative importance sampling procedure that satisfies both the norm preserving the condition and noise controlling condition in Definition~\ref{def:procedure_agnostic_learning-NN}.
\end{lemma}
\begin{proof}
Because the coefficient 
\begin{align*}
    \beta_i=1/k=O(\eps/K_{\alpha, D'})
\end{align*}
and 
\begin{align*}
    \sum_{i=1}^k \beta_i=1,
\end{align*}
this indicates the noise controlling condition holds.

Since 
\begin{align*}
    k = \Theta(K_{\alpha, D'} \log {d}),
\end{align*}
by Lemma~\ref{lmm:general_distribution-NN}, we know all eigenvalues of $A^* \cdot A$ are in $[1-1/2,1+1/2]$ with probability $1-10^{-3}$. 

This indicates the norm preserving condition holds.
\end{proof}

\section{Results for Function Family of Nearly Orthogonal Basis}
\label{sec:main_nearly_app}
Previous work ~\cite{cp19} only considers orthogonal basis, we generalize it into $\rho$-nearly orthogonal basis.

\begin{theorem}[Formal version of Theorem~\ref{thm:main_nearly_ortho}]
\label{cor:active_learning-NN}

If the following conditions hold
\begin{itemize}
    \item Suppose $\FF$ is a family of functions mapping from the domain $D$ to the set of real numbers $\R$, where the functions have a dimension of $d$. 
    \item Let $(D, Y)$ be an unknown distribution which is over $(x, y)$ in $\R^d \times \R$.
    \item The condition number is defined as 
  \begin{align}\label{eq:defineK}
    K := \underset{h \in \FF: h \neq 0}{\sup} \frac{\sup_{x \in D}|h(x)|^2}{\|h\|_D^2}.
  \end{align}
  \item Let $f^* \in \FF$ minimize $\E[\abs{f(x) - y}^2]$. 
  \item Let $\eps \in (0, 1/10)$.
\end{itemize}
 
 Then, there exists a randomized algorithm that takes
    \begin{align*}
        O((1+\rho d)(K\log (d )+ {K}/{\eps}))
    \end{align*}
    unlabeled samples from $D$ and requires $O({d}/{\eps})$ labels to output $\wt{f}$ such that
  \begin{align*}
    \E_{\wt{f}}\E_{x \sim D}[|\wt{f}(x)-f^*(x)|^2] \leq O( \eps(1+\rho d) ) \cdot \E_{x, y}[|y - f^*(x)|^2].   
  \end{align*}
\end{theorem} 

\begin{proof}
$\|f\|_{D'}$ represents 
\begin{align}
    \sqrt{\underset{x \sim D'}{\E}[|f(x)|^2]}.
\end{align}

By Lemma~\ref{lem:agnostic_learning_single_distribution-NN} with $D$ and the property of $P$, with probability at least $1-2 \cdot 10^{-3}$, 
\begin{align}\label{eq:D_D_0-NN}
\|h\|^2_{D_0} \in [\frac{3}{4C_r},\frac{5}{4C_l}] \cdot \|h\|^2_{D}   
\text{ for every } h \in \FF.
\end{align} 
We condition on Eq.~\eqref{eq:D_D_0-NN} holds from now on.

For all $i \in [k_0]$, we consider $y_i$ as a random label assigned to $x_i$, which is from $Y(x_i)$, containing both the unlabeled samples from the algorithm and the labeled samples from the Step~\ref{step-main:ERM} of Algorithm~\ref{alg-main:ag_unknown-NN}.

Let $y_i$ denote a random label of $x_i$ from $Y(x_i)$ for each $i \in [k_0]$ including the unlabeled samples in the algorithm and the labeled samples in Step~\ref{step-main:ERM} of Algorithm~\ref{alg-main:ag_unknown-NN}. 

Let $f'$ be defined as 
\begin{align}\label{eq:def_f'}
f'=\argmin_{h \in \FF} \underset{x_i \sim D_0,y_i \sim Y(x_i)}{\E} \left[ |y_i-h(x_i)|^2 \right].
\end{align}

Using Eq.~\eqref{eq:D_D_0-NN}, Theorem~\ref{thm:guarantee_AL_procedure-NN}, and Lemma~\ref{lem:agnostic_learning_single_distribution-NN}, we have
\begin{align}\label{eq:f'-f}
\underset{(x_1,y_1),\dotsc,(x_{k_0},y_{k_0})}{\E}[\|f'-f\|^2_D] \le \eps \cdot \underset{(x,y)\sim (D,Y)}{\E}[|y-f(x)|^2].
\end{align}

Using Eq.~\eqref{eq:D_D_0-NN} and the guarantee of Procedure $P$, we have 
\begin{align}\label{eq:wtf_f}
\underset{P}{\E}[\|\wt{f}-f'\|^2_{D_0}] \leq \eps \cdot \underset{x \sim D_0}{\E} \left[ |y_i-f'(x_i)|^2 \right]
\end{align}
from the proof of Theorem~\ref{thm:guarantee_AL_procedure-NN}.

Next, we bound the right hand side 
\begin{align*}
    \underset{x_i \sim D_0}{\E} \left[ |y_i-f'(x_i)|^2 \right]
\end{align*}
by 
\begin{align*}
    \underset{(x,y) \sim (D,Y)}{\E} \left[ |y-f(x)|^2 \right]
\end{align*}
over the randomness of $(x_1,y_1),\dotsc,(x_{k_0},y_{k_0})$:
\begin{align}\label{eq:E_E_y-f}
& ~ \E_{(x_1,y_1),\dotsc,(x_{k_0},y_{k_0})} \left[ \E_{x_i \sim D_0} \left[ |y_i-f'(x_i)|^2 \right] \right] \notag\\
\leq & ~ \E_{(x_1,y_1),\dotsc,(x_{k_0},y_{k_0})} \left[2\E_{x_i \sim D_0} \left[ |y_i-f(x_i)|^2 \right] + 2 \|f-f'\|_{D_0}^2 \right] \notag\\
\leq & ~ 2 \E_{(x,y) \sim (D,Y)} \left[ |y-f(x)|^2 \right] + \frac{3}{C_l} \E_{(x_1,y_1),\dotsc,(x_{k_0},y_{k_0})}\big[\|f-f'\|_D^2\big], 
\end{align}
where the first step follows from triangle inequality and the second step follows from Eq.~\eqref{eq:D_D_0-NN}. 

Hence
\begin{align}\label{eq:ff'_Eyf_E}
& ~ \underset{(x_1,y_1),\dotsc,(x_{k_0},y_{k_0})}{\E}\big[\underset{P}{\E}[\|\wt{f}-f'\|^2_{D_0}] \big] \notag \\
\leq & ~ \eps \cdot \underset{(x_1,y_1),\dotsc,(x_{k_0},y_{k_0})}{\E}\big[\underset{x \sim D_0}{\E} \left[ |y_i-f'(x_i)|^2 \right]] \notag \\
\leq & ~ \eps \cdot (2 \E_{(x,y) \sim (D,Y)} \left[ |y-f(x)|^2 \right] + \frac{3}{C_l} \E_{(x_1,y_1),\dotsc,(x_{k_0},y_{k_0})}\big[\|f-f'\|_D^2\big]) \notag \\
\leq & ~ \eps \cdot (2 \E_{(x,y) \sim (D,Y)} \left[ |y-f(x)|^2 \right] + \frac{3}{C_l} \eps \cdot \underset{(x,y)\sim (D,Y)}{\E}[|y-f(x)|^2]\big]) \notag \\
\leq & ~ \eps(2+\frac{3\eps}{C_l}) \cdot \underset{(x,y) \sim (D,Y)}{\E} \left[ |y-f(x)|^2 \right] \notag \\
\leq & ~ 3 \underset{(x,y) \sim (D,Y)}{\E} \left[ |y-f(x)|^2 \right]
\end{align}
where the first step follows from Eq.~\eqref{eq:wtf_f}, the second step follows from Eq.~\eqref{eq:E_E_y-f}, the third step follows from Eq.~\eqref{eq:f'-f}, and the last step follows from simple algebra.

Then, using Markov's inequality to above equation (Eq.~\eqref{eq:ff'_Eyf_E}), we show that $\wt{f}$ of a good output of $P$ with distribution $D_0$ satisfying
\begin{align}\label{eq:ff'_Eyf}
    \|\wt{f}-f'\|^2_{D_0} \leq 30 \underset{(x,y) \sim (D,Y)}{\E} \left[ |y-f(x)|^2 \right] 
\end{align}
which has a high probability ($1-1/10$).

Thus, by rescaling $\eps$, we can get
\begin{align*}
\|\wt{f}-f\|^2_D 
\leq & ~ 2 \|\wt{f}-f'\|^2_D + 2 \|f'-f\|^2_D \\
\leq & ~  \frac{8C_r}{3}\|\wt{f}-f'\|^2_{D_0} + 2 \|f'-f\|^2_D \\
\leq & ~ 80C_r\underset{(x,y)\sim (D,Y)}{\E}[|y-f(x)|^2] + \frac{\eps}{4} \cdot \underset{(x,y)\sim (D,Y)}{\E}[|y-f(x)|^2]\\ 
\leq & ~ (80C_r + \epsilon) \cdot \underset{(x,y)\sim (D,Y)}{\E}[|y-f(x)|^2]\\
\leq & ~ 80(1 + \rho d + \epsilon) \cdot \underset{(x,y)\sim (D,Y)}{\E}[|y-f(x)|^2],
\end{align*}
where the first step follows from the triangle inequality, the second step follows from Eq.~\eqref{eq:D_D_0-NN}, the third step follows from Eq.~\eqref{eq:ff'_Eyf}, the fourth step follows from simple algebra, and the fifth step follows from $C_r \leq 1+\rho d$.

Moreover, we claim that 
\begin{align*}
\underset{h \in \FF: h \neq 0}{\sup} \frac{\sup_{x \in D}|h(x)|^2}{\|h\|_D^2} \leq C_r \underset{h \in \FF: h \neq 0}{\sup} \frac{\sup_{x \in D}|h(x)|^2}{\|\alpha(h)\|_2^2}
\end{align*}
where $C_r \leq 1+\rho d$.

By applying Lemma \ref{lemma:BSS-NN},  we complete our proof.
\end{proof}

\section{Results for Active Deep Learning}
\label{sec:active_DL}
In this section, we apply the active learning theory above to active deep learning.

\begin{definition}[Two layer neural network]
\label{def:NN}
We define two layer neural network as follows
\begin{align*}
    f_{\nn}(W, a, x) := \frac{1}{\sqrt{m}}\sum_{r=1}^{m}a_r \phi(w_r^\top x) \in \R
\end{align*}
where $x\in\R^{d}$ is the input, $w_r\in\R^{d}, r \in [m] $ is the weight vector of the first layer, $ W=[w_1,\cdots, w_m]\in\R^{{d}\times m}, a_r\in\R, r\in[m] $ is the output weight, $ a=[a_1,\cdots, a_m]^\top$ and $\phi(\cdot)$ is the non-linear activation function.

Here we consider only training the first layer $W$ with fixed $a$, so we also write 
\begin{align*}
    f_{\nn}(W, x) = f_{\nn}(W, a, x).
\end{align*}

Given the training data matrix 
\begin{align*}
    X=[x_1, \cdots, x_n]\in \R^{n\times {d}}
\end{align*}
and labels $Y=[y_1,\cdots,y_n] \in \R^n $, we denote 
\begin{align*}
    f_\nn(W, X)=[f_\nn(W, x_1), \cdots, f_\nn(W,x_n)]^\top\in\R^n.
\end{align*}

\end{definition}

We prove our main theorem for active deep learning:
\begin{claim}\label{clm:exists_basis}
If the activation $\phi$ in a neural network $f_\nn$ satisfies the following conditions, 
\begin{itemize}
    \item $\phi^{(10d+\log(1/\eps_0)/\log(d))}(x)$  exists and is continuous.
    \item $\phi^{(10d+\log(1/\eps_0)/\log(d))}(x)\leq 1,~x\in \R $.
\end{itemize}
then there exists  $\{v_1,\ldots,v_{\ov{d}}\}$ forms a fixed $\rho$-nearly orthonormal basis of $\FF_\nn$ where
\begin{align*}
    \ov{d} \leq \binom{10d+\log(1/\eps_0)/\log(d)}{d}.
\end{align*}
Furthermore, for any $ W\in\R^{{d}\times m}$, there exists $h\in\FF_\nn$ such that 
\begin{align*}
  \|h(x)-f_\nn(W, x)\|_D^2\le \eps .
\end{align*}

Besides, for any $ h\in\FF_\nn$, there exists $ W\in\R^{d\times m}$ such that 
\begin{align*}
    \|h(x)-f_\nn(W, x)\|_D^2\le \eps .
\end{align*}
\end{claim}
\begin{proof}
For any activation $\phi:\R\rightarrow\R$ and input 
\begin{align*}
    z_r=w_r^\top x\in\R,
\end{align*}
$f$ can be expanded by Taylor's theorem
\begin{align*}
\phi(z_r)=\phi(0)+\phi'(0)z_r+\frac{\phi''(0)}{2!}z_r^2+\cdots+\frac{\phi^{(k)}(0)}{k!}z_r^k+\frac{\phi^{(k+1)}(\xi)}{(k+1)!}z_r^{k+1}
\end{align*}
where $\xi\in [0, z_r]$.

It's natural to consider the $W$ bound, eg NTK regime. In the NTK regime, we have that with a high probability,
\begin{align*}
    z_r 
    = & ~ w_r^\top x  \\
    \leq & ~ \| w_r \|_{\infty} \cdot \| x \|_1 \\
    \leq & ~  \|x\|_1 \\
    \leq & ~ \sqrt{d}  \|x\|_2 \\
    \leq & ~ \sqrt{d},
\end{align*}
where the first step follows from $z_r=w_r^\top x$ (see the beginning of the proof), the second step follows from $\langle a,b\rangle \leq \| a \|_{\infty} \cdot \| b \|_1$, the third step follows from $\| w_r \|_{\infty} \leq 1$, the fourth step follows from $\| x \|_1 \leq \sqrt{d} \| x \|_2$, the third step follows from $\|x\|_2 \leq 1$.

We can claim that 
\begin{align*}
    \frac{\phi^{(k+1)}(\xi)}{(k+1)!}z_r^{k+1}
    \leq & ~ (\frac{e}{k+1}z_r)^{k+1} \\
    \leq & ~ \eps_0,
\end{align*}
where the first step follows from $\phi^{(k+1)}(\xi)\leq 1$ and the second step follows from 
\begin{align*}
    k\geq (e\sqrt{d})^{2}+\log(1/\eps_0)/\log(e\sqrt{d}).
\end{align*}

Besides, taking the first $k+1$ terms in Taylor's theorem. Our neural network $f_\nn$ can be seen as a polynomial with $d$ variable and at most $k $ degree. So, our neural network $f_\nn$ can be seen as a polynomial with at most $ \binom{k+d}{d}$ terms. 

As any polynomial can be orthonormal decompose, we have that
\begin{align*}
  \ov{d}
  \leq & ~ \binom{k+d}{d} \\
  \leq & ~ \binom{10d+\log(1/\eps_0)/\log(d)}{d},
\end{align*}
where the second step follows from $d\geq e$.
\end{proof}

\begin{claim}
    We will claim that polynomial, ReLU, Sigmoid, and Swish hold the condition mentioned in Claim \ref{clm:exists_basis}.
\end{claim}

\begin{theorem}[Formal version of Theorem~\ref{thm-main:main}]
\label{thm:main}
If the following conditions hold
\begin{itemize}
    \item Let $f(W, x)$ be as defined in Definition~\ref{def:NN}
    \item $(D,Y)$ is an unknown distribution on $(x, y)$ over $\R^d \times \R$.   
    \item $D$ is the marginal distribution over $x$.
    \item Suppose $D$ has
  \begin{align}
    K := \underset{W \in \R^{d\times m}: W \neq 0}{\sup} \frac{\sup_{x \in D}|f(W,x)|^2}{\|f(W, x)\|_D^2},
  \end{align}
    which is a bounded condition number.
    \item Let $W^* \in \R^{d\times m}$ minimize $\E[\abs{f(W, x) - y}^2]$. 
    \item There exists $d \geq 3,\eps_0 \in (0,1/10)$.
    \item Let $ 0< \eps \leq O( 1/\log^3(d) )$.
\end{itemize}

  Then, there exists a randomized algorithm $P$ that takes
  \begin{align*}
      O(K\log (d )+ {K}/{\eps})
  \end{align*}
    unlabeled samples from $D$ and requires $O({d}/{\eps})$ labels to output $\wt{W}\in\R^{d\times  m}$ such that
  \begin{align*}
    \E_{P}\E_{x \sim D}[|f(\wt{W},x)-f(W^*,x)|^2] \leq O(\eps_0) + O(\eps) \cdot \E_{x, y}[|y - f(W^*,x)|^2].
  \end{align*}
\end{theorem} 
\begin{proof}
By Claim \ref{clm:exists_basis}, we have that there exists $\{v_1,\ldots,v_{d}\}$ forms a fixed $\rho$-nearly orthonormal basis of $\FF$, such that there exists $ h^*$ satisfied that 
\begin{align}\label{eq:hf_leq_e0}
\|h^*(x)-f(W^*, x)\|_D^2 \leq & ~ \eps_0 .
\end{align}

By Theorem \ref{cor:active_learning-NN}, 
 there exists a randomized algorithm that takes
\begin{align*}
    O(K\log (d )+ {K}/{\eps})
\end{align*}
unlabeled samples from $D$ and
  requires $O({d}/{\eps})$ labels to output $\wt{h}$ such that
  \begin{align}\label{eq:hh_leq_eyh}
    \E_{\wt{h}}\E_{x \sim D}[|\wt{h}(x)-h^*(x)|^2] \leq \eps \cdot \E_{x, y}[|y - h^*(x)|^2].
\end{align}
  
By Claim \ref{clm:exists_basis}, there exists $\wt{W}$ such that 
\begin{align}\label{eq:wthf_leq_e0}
\|\wt{h}(x)-f(\wt{W}, x)\|_D^2\le&~ \eps_0 .
\end{align}

As a result, 
\begin{align*}
& ~ \E_{P}\E_{x \sim D}[|f(\wt{W},x)-f(W^*,x)|^2]\\
\leq & ~ \E_{P}\E_{x \sim D}[2|h^*(x)-f(W^*,x)|^2+2|f(\wt{W},x)-h^*(x)|^2]  \\
\leq & ~ \E_{P}\E_{x \sim D}[2|h^*(x)-f(W^*,x)|^2+4|\wt{h}(x)-h^*(x)|^2+4|\wt{h}(x)-f(\wt{W},x)|^2]  \\
\leq & ~ 2 \E_{P}\E_{x \sim D}[|h^*(x)-f(W^*,x)|^2] + 4 \E_{P}\E_{x \sim D}[|\wt{h}(x)-h^*(x)|^2] + 4 \E_{P}\E_{x \sim D}[|\wt{h}(x)-f(\wt{W},x)|^2]  \\
\leq & ~ 2 \epsilon_0 + 4 \eps \cdot \E_{x, y}[|y - h^*(x)|^2] + 4 \epsilon_0  \\
\leq & ~ 10 \eps_0 + 4\eps \cdot \E_{x, y}[2|y - f(W^*, x)|^2+2|f(W^*, x) - h^*(x)|^2]\\
\leq & ~ 10 \eps_0 + 20\eps \cdot \E_{x, y}[|y - f(W^*, x)|^2],
\end{align*}
where the first step follows from the triangle inequality, the second step follows from the triangle inequality, the third step follows from the linearity property of expectation, the fourth step follows from combining Eq.~\eqref{eq:hf_leq_e0}, Eq.~\eqref{eq:hh_leq_eyh}, and Eq.~\eqref{eq:wthf_leq_e0}, the fifth step follows from the triangle inequality, and the last step follows from $\E[ |f(W^*,x) - h^*(x)|^2 ] \leq \E[ | y -f(W^*,x)|^2 ]$.

Therefore, we complete the proof.
\end{proof}

\section{Experiments}

\label{sec:CAHouse_expriments}

We implement our main Algorithm \ref{alg-main:ag_unknown-NN}, \ref{alg:BSS-NN} in our main Theorem \ref{thm:main_nearly_ortho}, and evaluate its performance on linear regression tasks on both the real-life dataset and the synthetic dataset. We also explore and verify the influence of the parameter $\eps$ in Theorem \ref{thm:main_nearly_ortho}. The results show that our algorithm works well theoretically and practically. 

In Section~\ref{sec:real_life_dataset_experiments}, we present the experiment on the real-life dataset. In Section~\ref{sec:experiment_on_synthetic_dataset}, we introduce the experiment on the synthetic dataset.

\subsection{Experiments on Real-life Dataset}
\label{sec:real_life_dataset_experiments}

Here, we start to introduce the experiment on the real-life dataset.

\paragraph{Experimental setup}

We test our algorithm in the California housing dataset ~\cite{pace1997sparse}, which contains 20,640 samples \footnote{The dataset is available in scikit-learn datasets.}. Each sample in the dataset has 8 numerical features and 1 target, and the target is within the range between 0.15 and 5.00. We randomly select 20\% of the data as testing data, and the rest 80\% as unlabelled data.  
We use the root mean square error (RMSE) as our evaluation metric, where a lower RMSE indicates a better performance. Because of the randomness of sampling in our Algorithm \ref{alg:BSS-NN}, we choose 10 different random seeds and report the mean and the standard deviation of the RMSE score to demonstrate the robustness of the algorithm.

\paragraph{Approximation ratio $\eps$}

\begin{table}[!ht]
\caption{Comparing the performance of our method under different $\eps$. Note that the row `Full' means labeling all the unlabeled data (=16,512) and using them to train the model. Values in the table indicate mean $\pm$ standard deviation. }
    \label{tab:approxi}
    \centering
    \begin{tabular}{|c|c|c|c|}
    \hline
        {\bf Setting} & {\bf Selected Samples} & {\bf RMSE} \\\hline
        Full & 16,512 ($\pm$ 0.000) & 0.718 ($\pm$ 0.000)  \\\hline
        $\eps=1$ & 113 ($\pm$ 2.049) & 0.742 ($\pm$ 0.009) \\\hline
        $\eps=0.1$ & 16,512 ($\pm$ 0.000) & 0.718 ($\pm$ 0.002) \\\hline
        $\eps=0.01$ & 16,512 ($\pm$ 0.000) & 0.712 ($\pm$ 0.001)
        \\\hline
    \end{tabular}
    
\end{table}

We demonstrate the effectiveness of our model, by comparing the RMSE score of our model trained by selected data with the RMSE score of the standard linear regression model trained using all training data. We try different $\eps$ and analyze its influence on the RMSE score. Table \ref{tab:approxi} shows the performance of our method under different $\eps$. As we can see, with the $\eps$ decreases, the difference between the RMSE score of our model and the RMSE score of the full data trained model,  decreases correspondingly. It is also worthwhile mentioning that when $\eps=1$, our model selects only 113 samples and achieves comparable performance with the model using all the training data. When $\eps=0.01$, our model selects all the samples for training and achieves even better performance. This is because our model assigns different importance weights to the selected samples while the standard linear regression model treats the samples as equally important.  

\begin{figure}[!ht]
    \centering
    \includegraphics[width=0.8\textwidth]{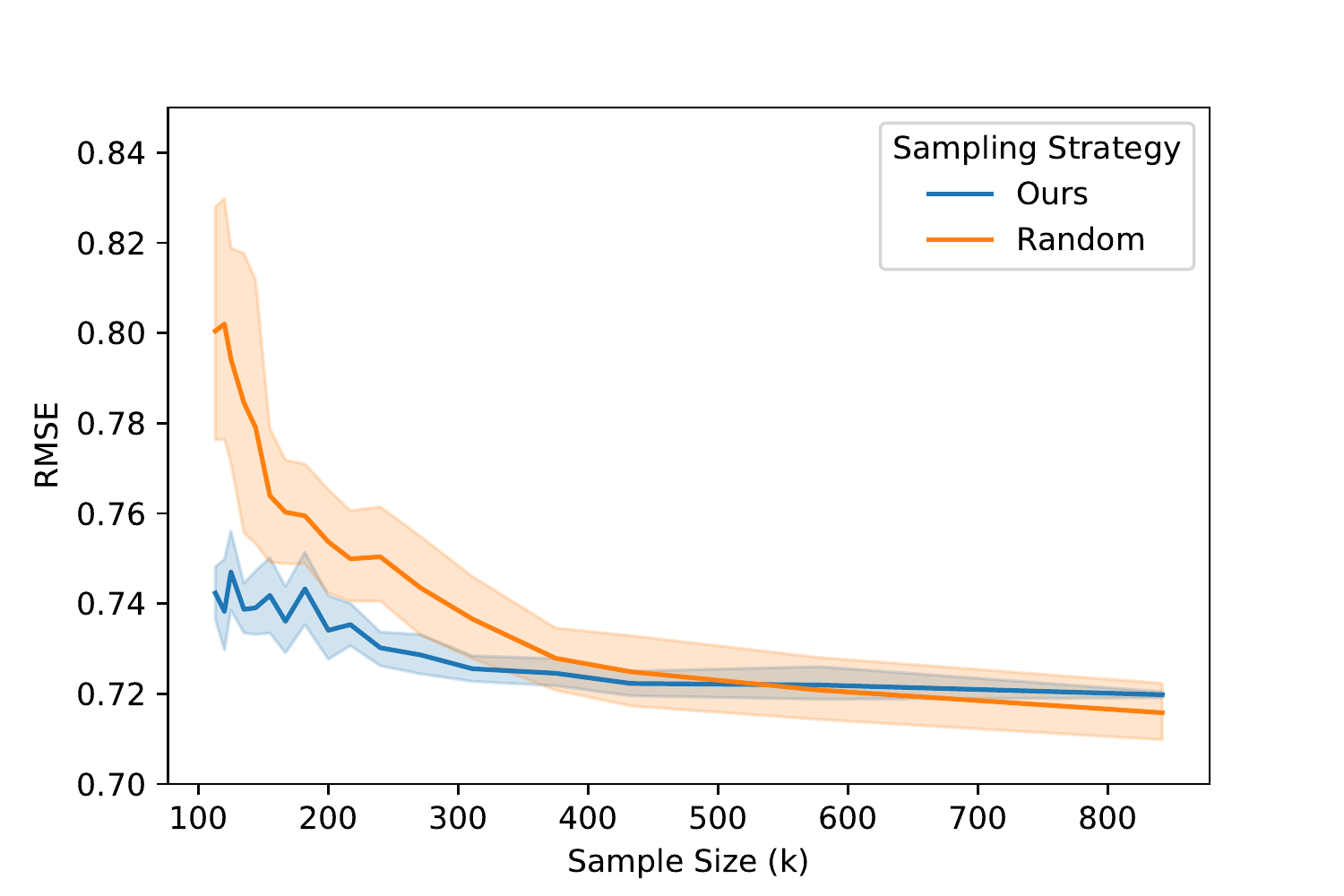}
    \caption{Comparing the performance of our sampling strategy with random sampling strategy under different sample size $k$.}
    \label{fig:CAHouse_sample_size_k}
\end{figure}

\paragraph{Sample size $k$} We also compare the performance of our importance sampling strategy with the random sampling strategy. When using a random sampling strategy, the samples are selected uniformly. However, our sampling strategy -- Algorithm \ref{alg:BSS-NN}, will select samples according to their importance scores. It should be noted that in Algorithm \ref{alg:BSS-NN}, the sample size $k$ is decided by the approximation ratio $\eps$. In our experiment, we first obtain the value of $k$ under different $\eps$, then select the same number of samples using the random sampling strategy. Figure \ref{fig:CAHouse_sample_size_k} compares the performance of our sampling strategy with the random sampling strategy. As we can see, when $k$ is small (e.g., $k \leq 400$), our sampling strategy outperforms the random sampling strategy. This is essential especially when the cost of data labeling of high and the sample size $k$ is limited.

\subsection{Experiments on synthetic dataset}
\label{sec:experiment_on_synthetic_dataset}

In this section, we start to introduce the experiment on the synthetic dataset.

\paragraph{Experimental setup} We also conduct experiments on a synthetic dataset for a linear regression task, which contains 30,000 samples. Each sample in this dataset has 10 numeric features and 1 target, and the value of the target is within the range between -7.24 and 6.43. In particular, we generate the dataset using the scikit-learn make\_regression() function, where the features are generated following a standard normal distribution, and the target is generated by firstly multiplying the feature matrix and a random weight matrix, then adding Gaussian noise to the target. Similar to the California house price prediction task in Section \ref{sec:real_life_dataset_experiments}, we split the dataset into two parts: 20\% for testing and 80\% for training. We run the experiments with 10 different random seeds to eliminate the impact of randomness. 

\begin{table}[!ht]
    \centering
    \begin{tabular}{|c|c|c|c|}
    \hline
        {\bf Setting} & {\bf Selected Samples} & {\bf RMSE} \\\hline
        Full & 24,000 ($\pm$ 0.00) & 0.507 ($\pm$ 0.000)  \\\hline
        $\eps=1$ & 139 ($\pm$ 1.72) & 0.530 ($\pm$ 0.008) \\\hline
        $\eps=0.1$ & 1,535 ($\pm$ 23.61) & 0.510 ($\pm$ 0.001) \\\hline
        $\eps=0.01$ & 16,187 ($\pm$ 2616.28) & 0.507 ($\pm$ 0.000)
        \\\hline
    \end{tabular}
    \caption{Comparing the performance of our method under different $\eps$ on synthetic dataset. 'Full' means labeling all the unlabeled data (=24,000) and using them to train the model. Values in the table indicate mean $\pm$ standard deviation. }
    \label{tab:approxi_synthetic_dataset}
\end{table}

\paragraph{Approximation ratio $\eps$} Similar to the findings on the California house price dataset, with $\eps$ decreases, the number of selected samples increases, and the RMSE score of our model get closer to the model trained with all the training data. 

\begin{figure}[!ht]
    \centering
    \includegraphics[width=0.8\linewidth]{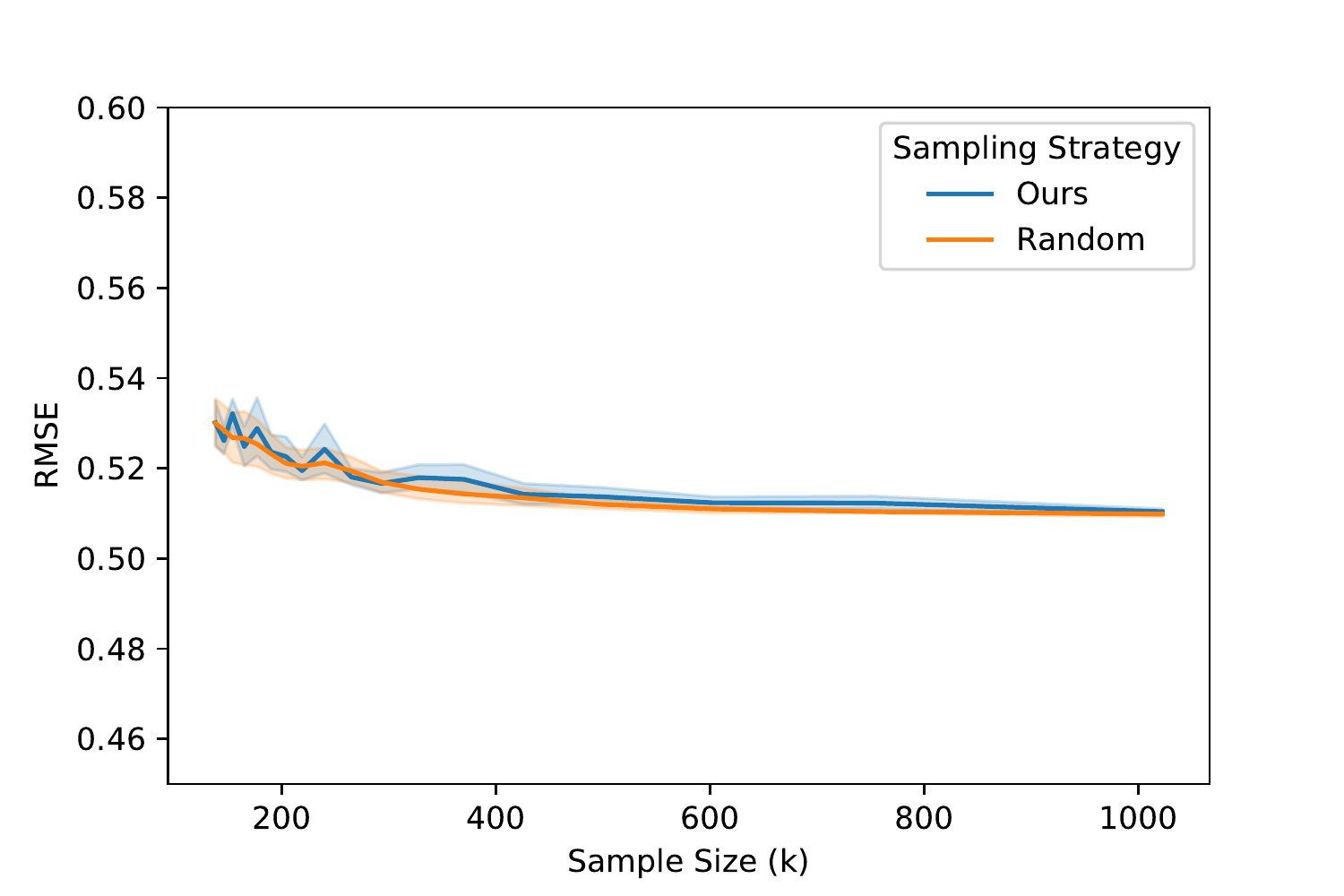}
    \caption{Comparing the performance of our sampling strategy with random sampling strategy under different sample size $k$ on the synthetic dataset. }
    \label{fig:synthetic_dataset_sample_size_k}
\end{figure}

\paragraph{Sample size $k$} Different from the findings on the California house price dataset, our importance sampling strategy achieves similar performance, compared to the random sampling strategy. This is because our importance sampling strategy relies on the feature distribution to select samples. When the features are randomly generated, the samples tend to have similar importance scores, which equals to the random sampling strategy. Even if our sampling strategy does not outperform the random sampling strategy on this synthetic dataset, this finding still validates the effectiveness of its performance.

\ifdefined\isarxiv
 \bibliographystyle{alpha}
 \bibliography{ref}

\else

\fi

\end{document}